\pdfoutput=1
\documentclass[11pt]{article}

\usepackage[preprint]{acl}

\usepackage{times}
\usepackage{latexsym}
\usepackage{amsmath}
\usepackage{bbold}
\usepackage{booktabs}
\usepackage{algorithm}
\usepackage{hyperref}

\usepackage[T1]{fontenc}
\usepackage[utf8]{inputenc}

\usepackage{microtype}

\usepackage{inconsolata}

\usepackage{graphicx}

\usepackage{multirow}

\newcommand{\wrongs}[1]{{\textcolor{red}{#1}}}
\newcommand{\ignore}[1]{{\textcolor{green}{#1}}}

\usepackage{makecell}
\title{Revisiting the Graph Reasoning Ability of Large Language Models: \\Case Studies in Translation, Connectivity and Shortest Path}

\author{Xinnan Dai$^1$, Qihao Wen$^1$, Yifei Shen$^2$, Hongzhi Wen$^1$,\\ \textbf{Dongsheng Li}$^2$, \textbf{Jiliang Tang}$^1$, \textbf{Caihua Shan}$^2$
\\Michigan State University$^1$, Microsoft Research$^2$\\\{daixinna, wenqihao, wenhongz, tangjili\}@msu.edu\\\{yifeishen, dongsli, caihuashan\}@microsoft.com}

\begin{document}
\maketitle
\begin{abstract}
Large Language Models (LLMs) have achieved great success in various reasoning tasks. However, their capacity for graph reasoning remains poorly understood. Although recent theoretical analyses suggest that LLMs can, in principle, perform complex graph tasks, empirical evaluations reveal numerous failures. To bridge this gap, we revisit the graph reasoning ability by introducing a new, balanced, and comprehensive benchmark. Through systematic experimentation, we identify key factors influencing performance, including node connectivity types, graph sizes, graph descriptions, and node naming methods. Moreover, we also demonstrate the impact of training data, model size and fine-tuning on graph reasoning. All the implementations and datasets are publicly available${\footnotemark[1]}$. 

% Meanwhile, we perform a real-world investigation on knowledge graphs and make consistent observations with our findings. The implementations and datasets are available${\footnotemark[1]}$. 
\end{abstract}
\footnotetext[1]{codes available: \url{ https://anonymous.4open.science/r/LLM-graph-evaluation-5E2C} \\datasets available: \url{https://drive.google.com/file/d/1tBQVW1ThflqdAV7iGW9oCpeHhwB8xghK/view?usp=sharing}.}
\section{Introduction}

Large Language Models (LLMs) have shown remarkable achievements in a multitude of reasoning tasks, ranging from mathematical, commonsense and symbolic problem-solving \cite{luo2023wizardmath, creswell2023selectioninference}, to more specialized applications like dialogue systems \cite{ouyang2022training}, program debugging \cite{surameery2023use} and scientific discovery \cite{boiko2023autonomous}. In this work, we focus on graph reasoning capability, where LLMs employ an explicit graph, sourced either from the input data or external resources, to infer the outcome. This reasoning ability is crucial and can be applied across various domains, such as   
improving question-answering system by a domain-specific knowledge graph \cite{huang2022language},
facilitating planning in autonomous agents through the tool relation graph \cite{liu2024toolnet}, and enhancing robot navigation via physical maps \cite{creswell2022selection}. 

%Previous studies explore the LLM's graph reasoning capability in a preliminary manner. GPT4Graph \cite{guo2023gpt4graph} and NLGraph \cite{wang2024can} proposed two benchmarks to evaluate the performance of LLMs in various graph-related tasks, showcasing their ability to tackle graph challenges.
%Furthermore, the works \cite{feng2024towards, wu2024can} proved that LLMs' proficiency in addressing dynamic programming problems theoretically. 

%To build a reliable reasoning path, these reasoning logics can be structured as an explicit graph-based reasoning framework to enhance the reasoning in question-answering \cite{huang2022language} and robotics agent planning \cite{creswell2022selection}. 
%Within this framework, graph connectivity is a fundamental tool to achieve complex reasoning in task planning tasks \cite{wu2024can} and knowledge reasoning tasks \cite{shu2024knowledge}. Meanwhile,  to ascertain the upper limits of LLM capabilities in graph reasoning, recent studies provide theoretical evidence that LLMs are adept at handling these tasks \cite{sanford2024understanding, de2024simulation}. Therefore, LLMs should have the potential to manage more complex challenges. 
\begin{figure}
    \centering
    \includegraphics[width=0.75\linewidth]{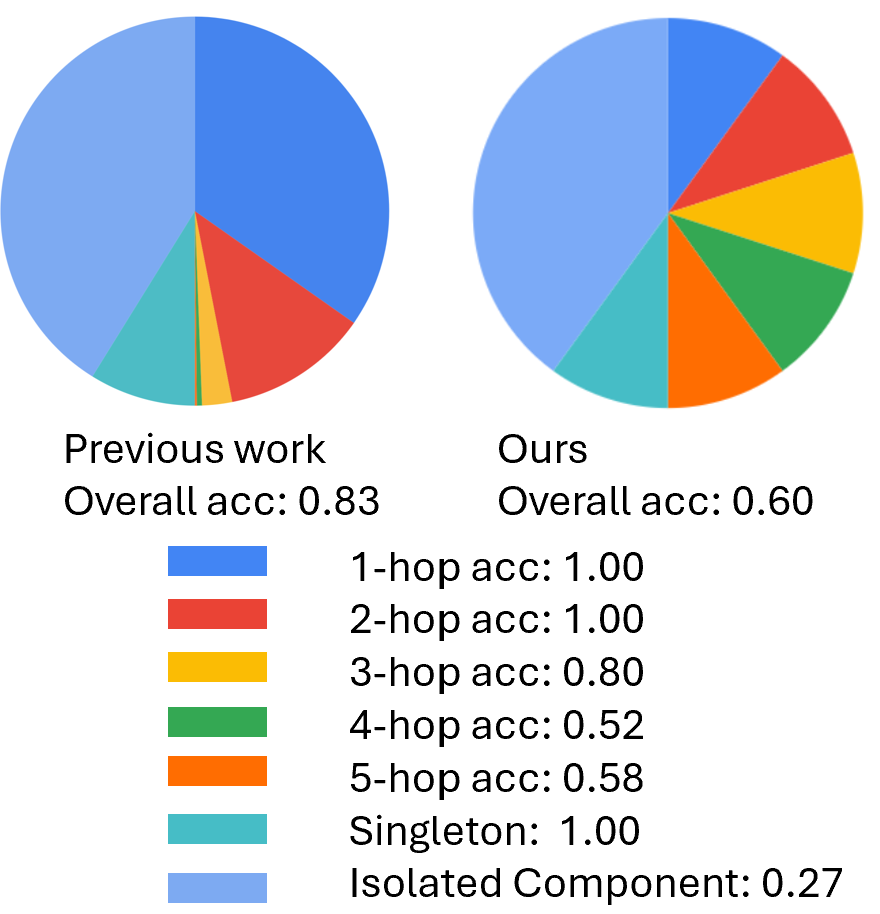}
    \caption{The overview of datasets in accuracy and distribution across different connectivity types. We evaluate GPT-3 on determining whether a path exists between two nodes. Previous work~\cite{wu2024can} primarily focused on 1-hop and 2-hop connections, resulting in high accuracy. However, it overlooked the fact that accuracy tends to drop when extending to 3, 4, and 5-hop connections.}
    \label{overview}
    \vspace{-0.25in}
\end{figure}

There are recent studies initially exploring the LLM's graph reasoning capability. On the one hand, the theoretical work \cite{feng2024towards} proved that LLMs have the ability to mimic a powerful decision-making framework (i.e., dynamic programming), to solve the complex tasks.  This suggests that LLMs are capable of handling certain graph reasoning tasks that can be formulated as decision-making problems, including breadth-first search for graph connectivity, and the Dijkstra for shortest path problem.
On the other hand, recent empirical studies, such as GPT4Graph \cite{guo2023gpt4graph} and NLGraph \cite{wang2024can}, found that LLMs could fail in these graph tasks. This discrepancy between theoretical expectations and practical observations indicates a critical gap in our comprehension of LLMs' graph reasoning abilities. 
In light of this, we aim to delve deeper into fundamental graph tasks to uncover the limitations inherent in LLMs, assess the impact of these limitations in real-world graphs, and propose possible explanations to understand the discrepancy.

In this work, we re-evaluate three fundamental graph reasoning tasks: graph description translation, graph connectivity, and the shortest path problem. First, we check whether LLMs can comprehend graph structures through the translation of varied graph descriptions (See Section~\ref{translation}). We summarize the three most popular graph description methods and evaluate the translation tasks among them. 
Although it is a simple reasoning task and 
LLMs could achieve high performance, LLMs are not entirely error-free. Then, we explore graph connectivity and examine LLMs systematically by considering varying connectivity lengths between nodes, diverse types of disconnections and different graph descriptions (See Section~\ref{connectivity}). Existing works~\cite{wang2024can, luo2024graphinstruct} primarily focus on the influence of graph size while considering only a limited range of connectivity types, leading to biased evaluations in connectivity tasks, as demonstrated in Figure~\ref{overview}. To address this, we construct a balanced and comprehensive dataset. Our investigations on this dataset indicate that in addition to graph size, node connectivity types and graph descriptions also play significant roles, and we extend those insights to the shortest path and real-world application tasks. In Section~\ref{sec:reasons}, we further explore the effects of model size and training data scale on graph reasoning, demonstrating that LLMs have the potential to excel in reasoning tasks given sufficient data and large-scale parameters. Finally, we reveal that LLMs may adopt different reasoning approaches depending on the form of the graph descriptions provided.

\section{Related work and Background}
% \label{naive_connect}
% In this section, we explore the theory that LLMs are capable of handling graph connectivity tasks. However, our experiments reveal that LLMs still still fail graph structure understanding, indicating a significant gap between theoretical predictions and experimental results in graph connectivity.

% Recent work shows that LLMs have an outstanding ability to perform graph reasoning tasks. Besides, the theoretical results suggest that LLMs are able to solve the graph reasoning tasks
\subsection{Evaluation on graph reasoning tasks}
Recent efforts have been made on graph reasoning evaluations \cite{guo2023gpt4graph, fatemi2023talk, mcleish2024benchmarking}. NLGraph \cite{wang2024can} evaluates LLMs across the 8 fundamental graph reasoning tasks, suggesting that LLMs have preliminary graph reasoning abilities. GraphInstruct \cite{luo2024graphinstruct} extends the graph reasoning benchmark to 21 classical graph tasks and introduces a step masking method to enhance the graph reasoning abilities of LLMs. Additionally, VisionGraph \cite{li2024visiongraph} provides a multimodal version of the graph reasoning task benchmark, extending its applicability beyond text.

\subsection{Graph connectivity in theory}
LLMs, through their transformer architecture, have demonstrated essential capabilities for reasoning tasks \cite{giannou2023looped, yang2023looped, sanford2024transformers}.
% The looped transformer architectures, \jt{Are the LLMs we evaluate in this work based on looped transformer?? } which is a widely adopted backbone of LLMs, have been proven to possess essential abilities for reasoning \cite{giannou2023looped, yang2023looped, sanford2024transformers}. 
Specifically, for the graph reasoning tasks, \citet{de2024simulation} suggest that looped transformers are able to simulate every step in a graph algorithm. \citet{sanford2024understanding} reveal that a single-layer transformer is sufficient for a naive graph connectivity task.

\subsection{LLMs for graphs in the applications}
Despite LLMs having capabilities in graph reasoning tasks in theory, there remains a gap between text understanding and graph reasoning \cite{chai2023graphllm,zhao2023graphtext}. Therefore, some recent work approves the use of additional tools to help LLMs understand graphs. Recent studies have validated the use of extra tools to enhance LLMs' comprehension of graphs. GraphEmb \cite{perozzi2024let} employs an encoding function to augment prompts with explicit structured information. Additionally, GraphWiz \cite{chen2024graphwiz} fine-tunes LLMs using graph reasoning datasets to achieve higher performance in graph tasks. However, when LLMs are pretrained using text data, their limitations in graph reasoning tasks remain unclear. In this work, we do a comprehensive study on the failures of LLMs in graph reasoning tasks. We summarize and analyze the potential reasons why LLMs fail in graph reasoning only using text prompts.

\subsection{Theoretical support for graph reasoning tasks}
\label{basic_theory}
\citet{feng2024towards} prove that if a task can be deconstructed into subtasks, it can be solved by LLMs. Based on this,  \citet{wu2024can} offer insights into transforming message-passing processes among graphs into subtasks of message-passing among nodes using transition functions, suggesting that LLMs are capable of handling graph decision tasks. Specifically, it can be theoretically proven that graph connectivity and shortest-path tasks are two examples of problems solvable by LLMs.
% Therefore, \jt{is the following a simple proof to the following statement?? if yes, we should have a separate subsection in this section which not only let readers know what tasks we focus in this work and why but also how LLMs can handle them theoretically. I think the reason we choose graph connectivity and shortest-path is that LLMs are theoretically proof to handle them but discrepancy existing in empirical evaluations. } graph connectivity tasks should be adapted to LLMs because they can also be deconstructed into subtasks.

Suppose that the structure of a graph can be represented as $\mathcal{G}=(\mathbf{X},\mathbf{E},\mathcal{E})$, where $\mathbf{X}$ is the set of nodes, $\mathbf{E}$ is the edge set, and $\mathcal{E}$ is the feature set of the edges. For the graph connectivity task, we start from node $n_i$ and end at node $n_j$. The transition function $F(i,j)$ for the graph connectivity task can be formulated as: $F(i,j)=\mathbb{1}_{k\in\mathcal{N}_{v_j}}(F(i-1,k)\cap F(k,j))$, where $\mathcal{N}_{v_j}$ denotes the neighbors of node node $v_j$ and $\mathbb{1}$ means whether the connection is existing. Consequently, we can deconstruct the graph connectivity tasks into sub-tasks, which have been proven to be solved by LLMs in \citet{feng2024towards}.

% \jt{is shortest path task a graph connectivity task??  }
% For the shortest-path task, we use the Bellman-Ford Algorithm as an example. This classical dynamic programming algorithm can be effectively solved by LLMs \cite{wu2024can}. The transition function is formulated as: $
% F(i,j)=\mathrm{min}_{k\in\mathcal{N}_{v_{j}}}(F(i-1,k)+\mathcal{E}_{kj})$, where $\mathcal{E}_{kj}$ is 1 for the unweighted graphs or the weight of the edge $E_{kj}$ in the weighted graphs.

% \jt{we always discuss graph connectivity but we also discuss shortest path. If shortest path is not a particuar type of connectivity task, we can summarize them as the fundemntal graph reasoning tasks which are theprocially proof to be handled by LLMs }
Theoretical results suggest that LLMs are capable of solving fundamental graph reasoning tasks, such as graph connectivity and shortest-path tasks. However, we find that they fail in practice.

\section{Limitations of LLMs in graph reasoning}
In this section, we empirically revisit the graph reasoning ability via case studies. In particular, we introduce three fundamental graph tasks: graph description translations in Section~\ref{translation}, graph connectivity in Section~\ref{connectivity}, and the shortest path task in Section~\ref{shortest-path_sec}. Finally, we summarize and analyze our findings in Section~\ref{analysis}.

% To revisit the ability of LLMs on the graph connectivity task,  we first categorize the types of connectivity and generate datasets based on these categories. Then, we do experiments to evaluate the effect of different connectivity types. %Further, we conclude the limitations of 

% The results presented in \ref{naive_connect} suggest that LLMs have a fundamental ability to handle the graph-related tasks. Despite this, LLMs perform well on basic connectivity tasks. In this section, to determine whether LLMs employ more advanced capabilities, such as reasoning, when addressing graph connectivity challenges, we reevaluate LLMs on the graph connective tasks.
\subsection{Graph description translation}

\label{translation}
% Graph descriptions are the foundational module to input graphs into LLMs.  Since LLMs are sensitive to input texts, varying graph descriptions can lead to different results. However, if LLMs understand graph structures, different descriptions should yield the same outcomes. Therefore, we set up an experiment to test this hypothesis.
\subsubsection{Graph Descriptions}
To begin with, we first describe the graph properties denoted as: {\it G describes a [attribute] graph among $ x\in\mathbf{X}$},
where [attribute] defines the graph types, such as undirected, directed, or knowledge graphs. Then, we use different graph descriptions to introduce their structures.

\begin{figure}
    \centering
    \includegraphics[width=0.5\textwidth]{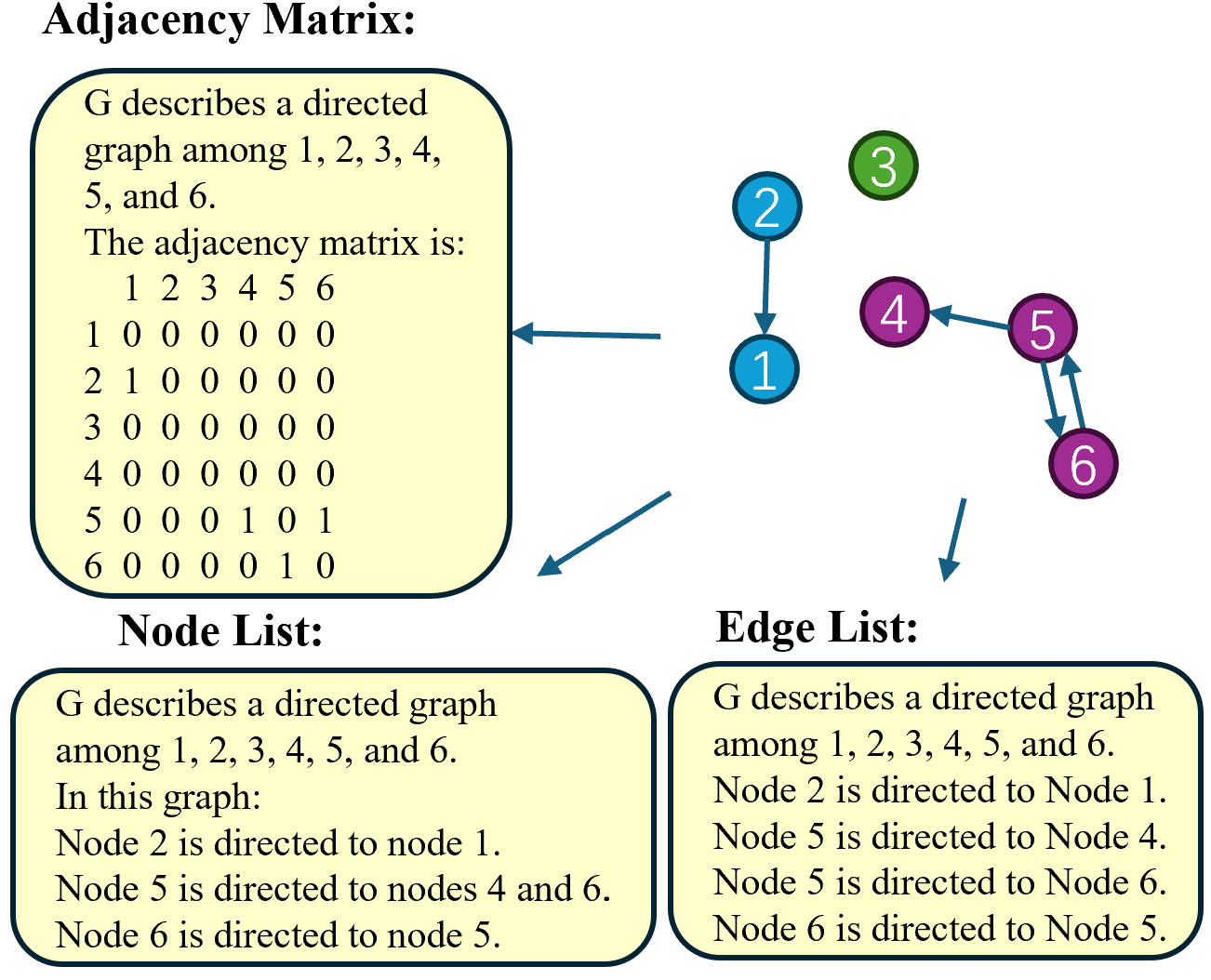}
    \caption{Three types of graph descriptions. A graph can be described by an adjacency matrix, edge list, and neighborhood node sets.}
    \label{fig-descript}
    \vspace{-0.2in}
\end{figure}

We summarize three types of graph structure description methods that have been widely used by the previous works \cite{fatemi2023talk,mcleish2024benchmarking} as shown in Figure~\ref{fig-descript}. They are (1) Adjacency Matrix: describing the adjacency matrix of a graph; (2) Node List: referring to the neighbors of a central node on a graph, and (3) Edge List: listing every edge of a graph. 
Adjacency Matrix is denoted as $\mathbf{A}\in\mathcal{R}^{N\times N}$, where $N$ is the number of nodes. In the text description, it encodes a paragraph by $N\times N$ binary tokens.

Node List uses the neighbors of a central node to describe a graph. For instance, consider the set of sentences $S_N = \{s_1, s_2, \dots, s_N\}$, which describes the graph via the neighbors $[u]$ of node $v_i$ with the edge feature $\epsilon$. A single sentence is as follows:% \jt{how to denote a node without any  neighbors???}
\begin{equation}
    s_i=\text{Node }{v_i} \text{ [relation] Nodes } \{[u,\epsilon]_{u\in \mathcal{N}_{v_i},\epsilon\in\mathcal{E}_(v_i,u)}\}. \nonumber
\end{equation}
Note that the [relation] varies across different types of graphs. In undirected graphs, we use the relation "is connected to," whereas in directed graphs, we use "is directed to." In knowledge graphs, the relation can be any specified type. 

Edge List describes a graph by listing the edges in a graph. The set of description sentences is denoted as: $S_{N_E}=\{s_1,s_2,\dots,s_{N_E}\}$, where $N_E$ is the number of edges and $s_i$ represents an edge, which is defined as:
\vspace{-0.1in}
\begin{equation}
    s_i=\text{Node } v_i \text{ [relation] Node } v_j, \epsilon_{ij}. \nonumber
\end{equation}
The examples of the graph descriptions are shown in Appendix~\ref{description_method}.
% \xn{Examples are in Appenix}

% \jt{the following can move to the next subsection since we did not even define the translation task yet}

% Therefore, in this paper, we focus on the widely used descriptions, Node and Edge.
% The descriptions of the graphs are the essential for LLMs

% If LLMs can accurately translate different descriptions of a graph, this suggests that they are capable of solving graph-related tasks by understanding graph structures. Otherwise, LLMs may need to adapt to alternative algorithms to effectively address these tasks.
\subsubsection{Translations on graph descriptions}

% \jt{we need to define the task??? what is graph description translation task??? I adjust the logic of the following paragraph, but need rewriting }

% If LLMs can understand the structures of a graph, such understandings should be independent on the ways to describe the gap. Therefore, to verify the ability of an LLM to understand the structural information of a graph, we design a xxxxxx task. It is .......  

If LLMs can comprehend the structures of a graph, such understanding should be independent of the methods used to describe the graph. Therefore, to verify the ability of LLMs to understand the structural information of a graph, we design a graph translation description task. This task requires LLMs to use the input graph description to generate various descriptions. After that, we will compare these descriptions to determine if they represent the same graph structure.
% For example,  the task of translating from Node List to Edge List typically involves first identifying the neighboring nodes of a specific node and then listing them. Conversely, translating from Edge List to Node List requires summarizing the information about the out-degree edges of a node and encapsulating it in a single sentence. This task is referred to as the "Neighborhood Prediction" task \cite{fatemi2023talk}.

Note that the number of tokens in the Adjacency Matrix depends on the number of nodes. This suggests that the Adjacency Matrix may require more tokens in dense graphs than Node or Edge Descriptions, limiting its applicability in the real world when the graph size is large.
Therefore, we only apply the Adjacency Matrix as the target format in the graph description translation task while employing Node List and Edge List as both source and target descriptions. Similarly, we use Node List and Edge List for graph connectivity and shortest-path tasks in Section ~\ref{connectivity} and Section ~\ref{shortest-path_sec}.

As suggested by the previous study, such as NLGraph~\cite{wu2024can} and GraphInstruct~\cite{luo2024graphinstruct}, increasing the graph size will challenge LLMs to understand graph structures. Thus, following the previous work, we use node numbers to indicate difficulty levels. In particular, we randomly generate 100 graphs with node numbers ranging from 5 to 25, and divide them into two datasets: one containing 50 graphs with node counts ranging from 5 to 15, labeled as "Easy", and another containing 50 graphs with node counts from 16 to 25, labeled as "Hard". 

We employ GPT-4 and LLAMA3.0-70B with the zero-shot setting and 0 temperature in the experiment. As the Adjacency Matrix is constrained by sentence length, we only predict the Adjacency Matrix on the dataset with smaller graphs. In the evaluation, we use the accuracy metrics. If the translations are completely correct, we categorize them as correct predictions. The results are summarized in the Table ~\ref{table1}.
\begin{table}[ht!]
    \centering
    \small
        \caption{Using LLMs to predict the translation among different descriptions. The scores are (GPT-4/LLAMA3.0-70B)}
    \label{table1}\setlength{\tabcolsep}{6pt}
\begin{tabular}{c|ccc}
\toprule
                             & \textbf{\# Graph}                          & \textbf{Avg. Node}                 & \textbf{Avg. Edge}                  \\
\multirow{-2}{*}{Dataset 1}  & \multicolumn{1}{c}{50}            & \multicolumn{1}{c}{10.6}  & \multicolumn{1}{c}{33.56}  \\\hline
\textbf{Source\textbackslash{}Target} & \textbf{Adjacency}                         &\textbf{ Nodes}                     & \textbf{Edges}                      \\
\textbf{Nodes}                        & 0.88 / 0.68                         & 1.00 / 0.94                    & 0.94 / 0.88                  \\
\textbf{Edges}                        & 0.88 / 0.66 & 0.94 / 0.74                 & 1.00 / 0.88                    \\\hline\hline
                             & \textbf{\# Graph}                          & \textbf{Avg. Node}                 & \textbf{Avg. Edge }                 \\
\multirow{-2}{*}{Dataset 2}  & \multicolumn{1}{c}{50}            & \multicolumn{1}{c}{20.49} & \multicolumn{1}{c}{110.35} \\\hline
\textbf{Source\textbackslash{}Target} & \textbf{Adjacency  }                       & \textbf{Nodes}                     & \textbf{Edges}                      \\
\textbf{Nodes }                       & -                                 & 1.00 / 0.90                    & 0.66 / 0.74                  \\
\textbf{Edges }                       & -                                 & 0.50 / 0.32                  & 0.92 / 0.70             \\
\bottomrule
\end{tabular}
\vspace{-0.2in}
\end{table}

The results indicate that LLMs struggle with graph description translations. LLMs achieve reliable accuracy only when the source and target descriptions are identical; however, they fail when translating between different types of descriptions. For example, LLMs show high accuracy in repeating the Node description, with both the source and target being Node descriptions.
However, their performance significantly declines when Edge Description is used. Similarly, while LLMs can summarize edge information effectively using Edge description, they struggle to summarize edge information from Node description. Those suggest that LLMs may not fully understand graph structures. %\jt{we may not use hallucinations, we may not fully understand the structure of a graph }hallucinations when using texts to build graph structures.

Furthermore, performance is also related to the sequence length. Although LLMs perform adequately with smaller-scale graphs, their effectiveness decreases as the graph size increases. Additionally, as Adjacency Matrix descriptions require more tokens in the output, accuracy significantly decreases when predicting adjacency matrices.  These findings align with similar limitations observed in general long-form text-generation tasks~\cite{ji2023survey}.

The experiments suggest that LLMs often generate content that is logically inconsistent with the input and the instructions, indicating that these failures may be due to faithfulness hallucinations.  The appendix~\ref{translation_failed} provides examples of these failures in description translation, where LLMs occasionally ignore certain edges or introduce non-existent ones, diverging from the input. Since translation tasks do not require complex reasoning but still exhibit hallucinations, it is possible that more complex reasoning tasks may also be prone to similar hallucinations in graph understanding.

% these errors are mainly from the mixture of training data. LLMs recall and blend graphs from their training data with the current input graph.

% the faithfulness hallucination causes failures in LLMs' understanding of graphs. 

% The experiments suggest that failure cases are possibly caused by faithfulness hallucination, i.e., LLMs tend to generate content logically inconsistent to the input and the instruction. in the tasks of graph description translation, LLMs occasionally ignore some edges or introduce non-existent edges, diverging from the input. \xn{shown in appendix} Since the translation tasks do not need complex reasoning, these errors are mainly from the mixture of training data. LLMs recall and blend graphs from their training data with the current input graph.

\begin{figure}
    \centering
    \includegraphics[width=0.3\textwidth]{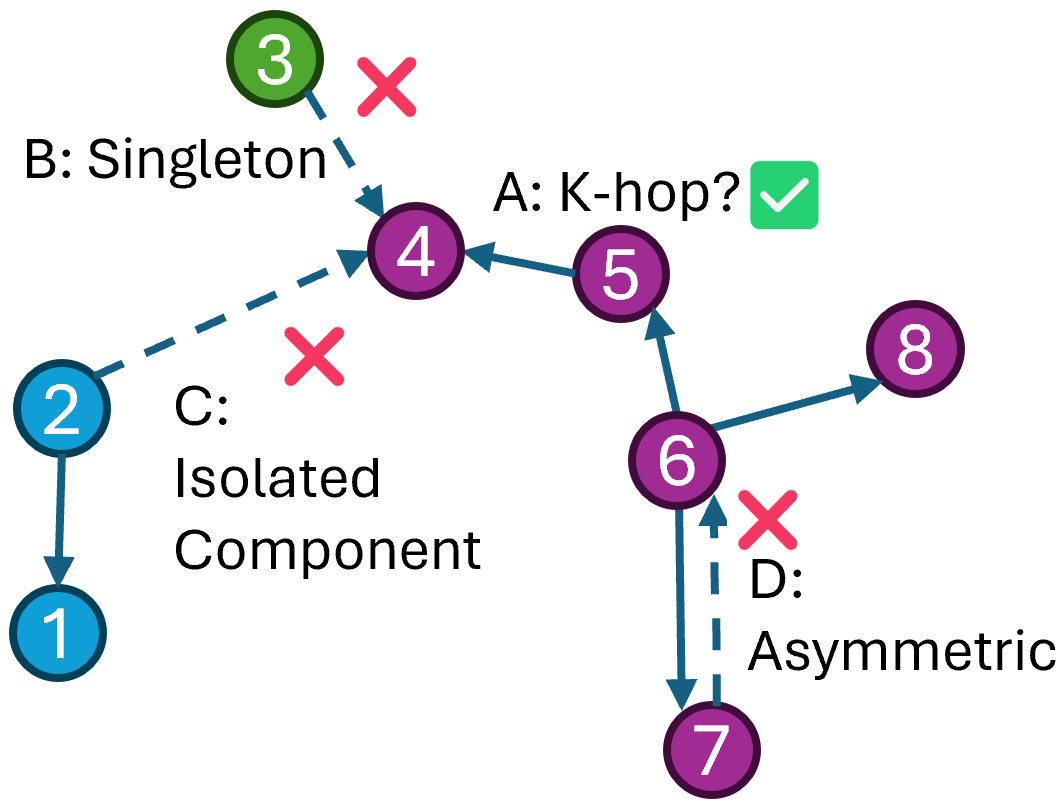}
    \caption{Different types of connectivity. The directed graph consists of 8 nodes, where solid lines represent the existence of directed edges, and dotted lines indicate no edge exists. Four connectivity types include: (A) K-hop: nodes 5 and 6 connect to node 4 within 1-hop and 2-hops, respectively. (B) Singleton: node 3 is an isolated node and not attached to node 4; (C) Isolated Components: nodes 2 and 4 belong to separate components with no path-connected edge; (D) Asymmetric: node 6 is directed towards node 7 but lacks any connection in an asymmetric configuration.}
    \label{fig_connect_types}
    \vspace{-0.2in}
\end{figure}

\begin{table*}[ht!]
% \vspace{}
    \centering
    \small
        \caption{Connectivity evaluation on the undirected graph datasets}
    \setlength{\tabcolsep}{4pt}
\resizebox{\textwidth}{!}{%
\begin{tabular}{c@{\hskip 0in}c@{\hskip 0in}c|ccccccccccccc}
\toprule
\multirow{2}{*}{ Difficulty } & \multirow{2}{*}{ Model } & \multirow{2}{*}{ Des. } & \multicolumn{3}{c}{k-hop, 1$\leq$k$\leq$2} & \multicolumn{3}{c}{k-hop, 3$\leq$k$\leq$4} & \multicolumn{3}{c}{5-hop} & Singleton & I.C. & \multirow{2}{*}{ AVG. ACC } & \multirow{2}{*}{ AVG. $\mathrm{F_{acc}}$ } \\
& & & ACC & $\mathrm{F_{acc}}$ & PCR & ACC & $\mathrm{F_{acc}}$ & PCR & ACC & $\mathrm{F_{acc}}$ & PCR & $\mathrm{F_{acc}}$ & $\mathrm{F_{acc}}$ & & \\\midrule
\multirow{4}{*}{ Easy } %& \multirow{2}{*}{ LLAMA3.0-8B } & Node & 0.98 & 0.88 & 0.85 & 0.97 & 0.69 & 0.91 & 0.96 & 0.46 & 0.95 & 0.72 & 0.11 & 0.60 & 0.48 \\
%& & Edge & 0.97 & 0.92 & 0.76 & 0.92 & 0.62 & 0.91 & 0.86 & 0.44 & 0.85 & 0.54 & 0.12 & 0.57 & 0.45 \\
& \multirow{2}{*}{ LLAMA3 } & Node & 1.00 & 0.99 & 0.99 & 1.00 & 0.96 & 0.98 & 1.00 & 0.92 & 0.96 & 1.00 & 0.33 & 0.73 & 0.71 \\
& & Edge & 1.00 & 0.94 & 0.88 & 1.00 & 0.96 & 0.98 & 0.98 & 0.78 & 0.94 & 1.00 & 0.44 & 0.77 & 0.73 \\
& \multirow{2}{*}{ GPT-3 } & Node & 1.00 & 0.98 & 0.82 & 0.88 & 0.87 & 0.93 & 0.78 & 0.72 & 0.87 & 0.92 & 0.13 & 0.60 & 0.59 \\
& & Edge & 1.00 & 0.96 & 0.80 & 0.82 & 0.80 & 0.93 & 0.88 & 0.72 & 0.90 & 0.94 & 0.17 & 0.61 & 0.58 \\
& \multirow{2}{*}{ GPT-4 } & Node & 1.00 & 0.93 & 0.99 & 1.00 & 0.93 & 0.99 & 1.00 & 0.94 & 0.97 & 1.00 & 0.53 & 0.81 & 0.78 \\
& & Edge & 1.00 & 0.93 & 0.98 & 1.00 & 0.90 & 0.98 & 0.98 & 0.88 & 0.97 & 0.98 & 0.69 & 0.87 & 0.83 \\\midrule
% \multirow{4}{*}{ Medium } & \multirow{2}{*}{ LLAMA3.0-8B } & Node & 0.98 & 0.89 & 0.78 & 0.96 & 0.66 & 0.95 & 1.00 & 0.44 & 0.92 & 0.56 & 0.08 & 0.58 & 0.44 \\
% & & Edge & 0.97 & 0.84 & 0.66 & 0.97 & 0.55 & 0.85 & 0.98 & 0.42 & 0.94 & 0.38 & 0.07 & 0.55 & 0.39 \\
\multirow{4}{*}{ Medium } & \multirow{2}{*}{ LLAMA3 } & Node & 1.00 & 0.94 & 0.90 & 1.00 & 0.93 & 0.96 & 0.94 & 0.82 & 0.93 & 1.00 & 0.36 & 0.74 & 0.70 \\
& & Edge & 1.00 & 0.96 & 0.83 & 0.96 & 0.81 & 0.90 & 0.94 & 0.62 & 0.94 & 0.98 & 0.35 & 0.72 & 0.65 \\
& \multirow{2}{*}{ GPT-3 } & Node & 1.00 & 0.97 & 0.72 & 0.81 & 0.74 & 0.84 & 0.76 & 0.62 & 0.79 & 0.94 & 0.16 & 0.60 & 0.56 \\
& & Edge & 1.00 & 0.96 & 0.72 & 0.72 & 0.60 & 0.90 & 0.76 & 0.52 & 0.83 & 0.96 & 0.18 & 0.59 & 0.53 \\
& \multirow{2}{*}{ GPT-4 } & Node & 1.00 & 0.89 & 0.98 & 1.00 & 0.85 & 0.97 & 1.00 & 0.94 & 0.92 & 0.98 & 0.42 & 0.77 & 0.71 \\
& & Edge & 1.00 & 0.91 & 0.97 & 1.00 & 0.90 & 0.93 & 0.96 & 0.74 & 0.94 & 0.96 & 0.44 & 0.77 & 0.71 \\\midrule
\multirow{4}{*}{ Hard } & \multirow{2}{*}{ LLAMA3 } & Node & 1.00 & 0.98 & 0.90 & 1.00 & 0.83 & 0.94 & 0.96 & 0.64 & 0.94 & 0.96 & 0.2 & 0.67 & 0.60 \\
& & Edge & 1.00 & 0.92 & 0.78 & 0.92 & 0.59 & 0.86 & 0.94 & 0.42 & 0.92 & 0.84 & 0.2 & 0.64 & 0.51 \\
& \multirow{2}{*}{ GPT-3 } & Node & 1.00 & 0.92 & 0.65 & 0.76 & 0.67 & 0.85 & 0.80 & 0.50 & 0.77 & 0.98 & 0.14 & 0.59 & 0.52 \\
& & Edge & 1.00 & 0.92 & 0.66 & 0.65 & 0.47 & 0.86 & 0.74 & 0.38 & 0.81 & 1.00 & 0.18 & 0.58 & 0.49 \\
& \multirow{2}{*}{ GPT-4 } & Node & 1.00 & 0.87 & 0.98 & 0.99 & 0.84 & 0.93 & 0.98 & 0.76 & 0.90 & 1.00 & 0.30 & 0.72 & 0.64 \\
& & Edge & 1.00 & 0.86 & 0.94 & 0.93 & 0.69 & 0.87 & 0.90 & 0.58 & 0.90 & 0.92 & 0.34 & 0.71 & 0.60 \\
\bottomrule
\end{tabular}
}
    \label{undirected_results}
    \vspace{-0.1in}
\end{table*}

\section{Revisit graph connectivity task}
\label{connectivity}
\subsection{Connectivity types}
Previous studies suggest that large language models (LLMs) possess essential capabilities for graph connectivity tasks \cite{wang2024can, luo2024graphinstruct}, yet they still fail in some instances. To further investigate the graph connectivity task, we begin by analyzing the samples where failures occurred based on those two baseline datasets.

We first categorize the types of connectivity samples. For the samples of connected nodes, we classify them according to the path length, which is denoted as K-hops. Besides, for the samples of unconnected nodes, we label them into three categories: Singleton, Isolated Components (IC), and Asymmetric, as shown in Figure~\ref{fig_connect_types}. Singleton denotes that one node is isolated. Isolated Components indicate that these two nodes belong to separate components in the graph. Note that a Singleton is a special case of Isolated Components. The distinction lies in the representations using Node List and Edge List, where the isolated node is not included in the descriptions of the graph structure, such as Node 3 in Figure~\ref{fig-descript}. Asymmetric is designated for directed graphs, highlighting situations where a path exists from one node to another, but the reverse path does not exist, indicating a one-way connectivity.

We calculate the distribution of connectivity types in the baseline datasets,  as shown in Table~\ref{all_dataset} of Appendix~\ref{graph_connect_datasets}, and subsequently conduct the experiment on them. The results, presented in Table~\ref{baseline_all} of Appendix~\ref{graph_connect_results}, indicate that the baseline datasets lack a balanced distribution across different connectivity types. More importantly, LLMs exhibit varying performances across these types. Thus, it is crucial to establish a balanced dataset to better evaluate different graph connectivities.

\begin{table*}[]
\vspace{-1cm}
    \centering
    \small
        \caption{ Connectivity evaluation on the directed graph datasets}
    \setlength{\tabcolsep}{3pt}
\resizebox{\textwidth}{!}{% 
\begin{tabular}{c@{\hskip 0in}c@{\hskip 0in}c|cccccccccccccc}
\toprule
\multirow{2}{*}{ Difficulty } & \multirow{2}{*}{ Model } & \multirow{2}{*}{ Des. } & \multicolumn{3}{c}{k-hop, 1$\leq$k$\leq$2} & \multicolumn{3}{c}{k-hop, 3$\leq$k$\leq$4} & \multicolumn{3}{c}{5-hop} & Singleton & Isolated C. & Asymmetric & \multirow{2}{*}{ AVG. ACC } & \multirow{2}{*}{ AVG. $\mathrm{F_{acc}}$ } \\
& & & ACC & $\mathrm{F_{acc}}$ & PCR & ACC & $\mathrm{F_{acc}}$ & PCR & ACC & $\mathrm{F_{acc}}$ & PCR & $\mathrm{F_{acc}}$ & $\mathrm{F_{acc}}$ & $\mathrm{F_{acc}}$ & & \\
\midrule
\multirow{4}{*}{ Easy } & \multirow{2}{*}{ GPT-3 } & Node & 0.99 & 0.92 & 0.88 & 0.85 & 0.58 & 0.93 & 0.92 & 0.36 & 0.95 & 0.96 & 0.13 & 0.37 & 0.66 & 0.53 \\
& & Edge & 1.00 & 0.93 & 0.94 & 0.89 & 0.47 & 0.95 & 0.92 & 0.30 & 0.96 & 0.94 & 0.15 & 0.36 & 0.67 & 0.51 \\
& \multirow{2}{*}{ GPT-4 } & Node & 0.99 & 0.98 & 0.94 & 0.95 & 0.81 & 0.96 & 0.88 & 0.66 & 0.96 & 1.00 & 0.84 & 0.85 & 0.91 & 0.86 \\
& & Edge & 0.99 & 0.97 & 0.99 & 0.88 & 0.72 & 0.97 & 0.76 & 0.44 & 0.99 & 0.98 & 0.65 & 0.84 & 0.85 & 0.78 \\
\midrule
\multirow{4}{*}{ Medium } & \multirow{2}{*}{ GPT-3 } & Node & 1.00 & 0.87 & 0.67 & 0.81 & 0.40 & 0.75 & 0.78 & 0.38 & 0.88 & 1.00 & 0.17 & 0.48 & 0.67 & 0.52 \\
& & Edge & 0.99 & 0.84 & 0.80 & 0.79 & 0.30 & 0.90 & 0.78 & 0.32 & 0.97 & 1.00 & 0.18 & 0.42 & 0.65 & 0.48 \\
& \multirow{2}{*}{ GPT-4 } & Node & 1.00 & 0.94 & 0.95 & 0.86 & 0.55 & 0.94 & 0.74 & 0.50 & 0.82 & 1.00 & 0.70 & 0.67 & 0.82 & 0.72 \\
& & Edge & 0.98 & 0.88 & 0.96 & 0.79 & 0.43 & 0.92 & 0.70 & 0.38 & 0.91 & 1.00 & 0.53 & 0.75 & 0.78 & 0.66 \\
\midrule
\multirow{4}{*}{ Hard } & \multirow{2}{*}{ GPT-3 } & Node & 0.98 & 0.81 & 0.53 & 0.65 & 0.25 & 0.71 & 0.80 & 0.26 & 0.77 & 1.00 & 0.10 & 0.55 & 0.64 & 0.47 \\
& & Edge & 0.93 & 0.75 & 0.74 & 0.64 & 0.19 & 0.86 & 0.84 & 0.16 & 0.89 & 0.98 & 0.18 & 0.57 & 0.65 & 0.45 \\
& \multirow{2}{*}{ GPT-4 } & Node & 0.96 & 0.88 & 0.91 & 0.81 & 0.44 & 0.81 & 0.68 & 0.36 & 0.76 & 0.98 & 0.70 & 0.53 & 0.77 & 0.64 \\
& & Edge & 0.96 & 0.80 & 0.93 & 0.85 & 0.40 & 0.83 & 0.76 & 0.38 & 0.82 & 0.98 & 0.41 & 0.59 & 0.74 & 0.58 \\
\bottomrule
\end{tabular}
}

    %Results on the directed graph datasets. Note that due to input length limitations, GPT-3 with Edge descriptions for the Hard subset could not deal with 15, 2, 5, and 3 samples in the subsets of 1<k<2 hops, 3<k<4 hops, path-connected (P.C), and Asymmetric subsets, respectively.}
    \label{directed_graphs}
    \vspace{-0.1in}
\end{table*}
\subsection{Dataset Construction}
In previous work, NLGraph~\cite{wang2024can} included only an undirected graph dataset for the connectivity task, and GraphInstruct~\cite{luo2024graphinstruct} featured an unbalanced distribution as shown in Appendix~\ref{graph_connect_datasets}, Table~\ref{all_dataset}. Therefore, based on these studies, we need to consider factors such as the number of nodes in graphs, edge directions, and types of connectivity.

Following previous studies~\cite{wu2024can,luo2024graphinstruct}, we indicate the difficulty levels of graphs based on the number of nodes, labeling them as Easy, Medium, and Hard. For each level, we initially generate all possible graphs with a certain number of nodes and then randomly select graphs and corresponding node pairs to formulate test pairs of the questions. For samples connected within K-hops, we collect 50 samples for each $k$ where $k\in[1,2,3,4,5]$. For negative samples, we selected 200 Isolated Component samples and 50 Singleton samples from the undirected graph dataset. Similarly, for the directed graph dataset, we chose 100 Connected pairs, 100 Asymmetric samples, and 50 Singleton samples. The details can be found in Table~\ref{all_dataset} of Appendix~\ref{graph_connect_datasets}. 

% We gather 50 samples where at least one node is isolated, creating the Singleton subset. \jt{how about "Isolated Components"???} To \jt{I do not get the following mean???} balance positive and negative samples in the dataset, we select 200 samples for the undirected graph dataset.

% \jt{this part should put in the beinning of this subsection to show their limitations and we need our dataset}

\subsection{Evaluation Metrics}
Instead of only evaluating the accuracy of graph connectivity, we also want to check if the reasoning path to make the prediction can support the prediction. Thus, the prompt is defined as follows: "If a path exists, present the path formatted as "Node \#1 -> Node \#2."; If no path is found, state "No path.". Therefore, to evaluate the reliability of such paths, we design two novel metrics, $\mathrm{Fidelity}_{\text{Acc}}$ ($\mathrm{F_{acc}}$) and Path Consistency Ratio (PCR), which are used to analyze the correctness of reasoning paths. $\mathrm{F_{acc}}$ evaluates whether the reasoning path to infer the answer is correct or not. The formulation is denoted as: $\mathrm{F_{acc}}=\frac{1}{M} \sum_{i=1}^M\left(\hat{y}_i=y_i\right)\wedge\left(\hat{p_i}\in \mathbf{P}\right)$,
where $\hat{y}_i$ denotes the predicted answer, $y_i$ the ground truth answer, $\hat{p}_i$ the predicted path, and $\mathbf{P}$ the set of reachable paths. $M$ is the number of data samples. $\mathrm{F_{acc}}$ correctly identifies the answer only when both the connective prediction and the path prediction are accurate. The range of $\mathrm{F_{acc}}$ is $\left[0, 1\right]$, where a higher score indicates greater consistency with the ground truth. A high accuracy with a low $\mathrm{F_{acc}}$ score suggests that the reasoning paths cannot well support connectivity predictions, which could  indicate that LLMs are hallucinating.

% Further, to evaluate whether LLMs have \jt{we need a more detail description about the purpose for this metric??? do we want to path as short as possible to indicate teh capacity of LLMs???? } 
Multiple reachable paths exist within a graph. LLMs demonstrate superior reasoning abilities if they can identify a shorter path. To assess the paths LLMs select for reasoning, we introduce the Path Consistency Ratio (PCR): $\mathrm{PCR}=\frac{1}{M} \sum_{i=1}^M\frac{|{p_i}|}{|\hat{p_i}|}$,
$|\hat{p}_i|$ represents the number of nodes in the path, while $|p_i|$ denotes the number of nodes in the shortest path. 
%The PCR score ranges from $PCR \in [0, 1]$. 
We evaluate PCR only when the LLMs give the correct path. A higher score indicates that the LLMs are more adept at selecting the shortest path between two nodes.

\subsection{Results}
\label{sec:connect_results}

We select three representative large language models, GPT-3 ( GPT-3.5-turbo-0301), GPT-4 (GPT-4-0125-preview) and LLAMA 3 (LLAMA3.0-70B). 
\paragraph{Undirected Graph Results}
We start with the undirected graph datasets and show the results in Table \ref{undirected_results}. 
First of all, GPT-4 has better reasoning ability compared with GPT-3 and LLAMA 3 across all cases, regardless of the graph difficulty, graph description or the categories of connectivity. %Additionally, LLAMA3-70B performs much better than LLAMA3-8B, indicating that larger LLMs have a better understanding of graphs.
%The observations on the different connectivity types are as follows: 

Secondly, we have following observations by comparing different connectivity situations:
%(1) $\mathrm{F_{acc}}$ decreases as the number of k-hops and graph nodes increases, highlighting these factors as crucial in defining tasks related to graph connectivity. 
(1) The difficulty of reasoning increases as the path length extends (i.e., K-hop), peaking in the isolated component (where K can be viewed as infinite). As a result, both ACC and $\mathrm{F_{acc}}$ exhibit a corresponding decline.
(2) The value of PCR is stable and almost larger than 0.9 via GPT-4, indicating a tendency of GPT-4 to find some shorter paths when judging the connectivity.
(3) The Singleton scene is particular because it is not affected by the difficulty changes and always performs well. This suggests that LLMs may have a shortcut in graph understanding: nodes not mentioned in the graph description are considered isolated and no connection with others. (4) Node Lists generally perform better than Edge Lists in most cases. This is because the search space differs when various description methods are used to search nodes within the next-token prediction framework.  For the Node Lists, it is easy to find all the positions of neighbor nodes, which costs $\mathcal{O}(|N|)$. However, it takes  $\mathcal{O}(|E|)$ for Edge Lists. Therefore, the overall algorithmic complexity is different, where the Node Lists should be $\mathcal{O}(|N|^2)$ while the Edge Lists should be $\mathcal{O}(|N||E|)$.

Interestingly, LLMs demonstrate enhanced performance with node descriptions when $k$ is larger, e.g., 5-hops, while they perform better in the isolated component scene when provided with edge descriptions. This suggests that LLMs may not consistently apply the same strategy for analyzing graph connectivity. Instead, the approach adopted by LLMs is shaped by the input context provided.

\paragraph{Directed Graph Results}

Next, we evaluate the connectivity on the directed graphs shown in Table \ref{directed_graphs}. Some key observations are similar to those of undirected graph datasets. However, LLMs have lower performance on directed graphs across almost all sub-datasets, yet they maintain high performance on subsets with k$\leq$2 and Singleton subsets.

% \xn{issue in the table data}

We also note distinct performance differences between GPT-3 and GPT-4 on the Asymmetric dataset. GPT-3's accuracy increased from 0.4 to 0.55, whereas GPT-4's decreased from 0.8 to 0.55. Given that an accuracy of 0.55 is nearly equivalent to random guessing in a binary task for asymmetric detection, it suggests that LLMs might engage in random reasoning within the Hard dataset. Furthermore, descriptions using Node Lists outperform those using Edge Lists. Since an Edge List simply describes two nodes in one sentence, LLMs may meet hallucination in determining whether the relationship "A is B" is equivalent to "B is A" \cite{berglund2023reversal}.
% We also observe that performance on the Asymmetric dataset would decrease as the difficulty increases.
% This means that LLMs are not sensitive to the sequence orders, aligning with \xn{cite}. Node descriptions show slightly better performances than Edge descriptions. \xn{Reason}. The reason is

% \begin{table}[]
%     \centering
%     \tiny
%     \begin{tabular}{cccccc}
% \multicolumn{2}{c}{Dataset} & \multicolumn{2}{c}{undirected graphs} & \multicolumn{2}{c}{directed graphs} \\
% Subdataset & Des. & unweighted & weighted & unweighted & weighted \\
% \multirow{2}{*}{ 1<k<2 hops } & Node & 0.88 & 0.79 & 0 & 0.72 \\
% & Edge & 0.89 & 0.68 & 0 & 0.65 \\
% \multirow{2}{*}{ 3<k<4 hops } & Node & 0.87 & 0.5 & 0 & 0.43 \\
% & Edge & 0.84 & 0.44 & 0 & 0.34 \\
% \multirow{2}{*}{ 5-hops } & Node & 0.88 & 0.54 & 0 & 0.38 \\
% & Edge & 0.68 & 0.42 & 0 & 0.22 \\
% \multirow{2}{*}{ attached } & Node & 1 & 0.98 & 0 & 0.96 \\
% & Edge & 1 & 0.98 & 0 & 0.96 \\
% \multirow{2}{*}{ P.C. } & Node & 0.665 & 0.745 & 0 & 0.94 \\
% & Edge & 0.81 & 0.634 & 0 & 0.89 \\
% \multirow{2}{*}{ Asymmetric } & Node & - & - & 0 & 0.78 \\
% & Edge & - & - & 0 & 0.72 \\
% \multirow{2}{*}{ AVG } & Node & 0.804 & 0.708 & 0 & 0.708 \\
% & Edge & 0.838 & 0.6176 & 0 & 0.638 \\
% \end{tabular}
%     \caption{Shortest path results}
%     \label{shortest-path}
% \end{table}

\begin{table}[]
\vspace{-1cm}
    \centering
    %\small
        \caption{Results on the shortest path problem}
\resizebox{0.98\columnwidth}{!}{%
\begin{tabular}{cc|cc|cc}
\toprule
\multicolumn{2}{c}{Dataset} & \multicolumn{2}{|c|}{undirected graphs} & \multicolumn{2}{c}{directed graphs} \\
Subdataset & Des. & unweighted & weighted & unweighted & weighted \\
\midrule
\multirow{2}{*}{ 1$\leq$k$\leq$2 hops } & Node & 0.88 & 0.80 & 0.93 & 0.76 \\
& Edge & 0.89 & 0.70 & 0.91 & 0.71 \\
\multirow{2}{*}{ 3$\leq$k$\leq$4 hops } & Node & 0.87 & 0.52 & 0.64 & 0.45 \\
& Edge & 0.81 & 0.47 & 0.51 & 0.38 \\
\multirow{2}{*}{ 5-hops } & Node & 0.88 & 0.54 & 0.48 & 0.40 \\
& Edge & 0.76 & 0.44 & 0.42 & 0.26 \\
\multirow{2}{*}{ Singleton } & Node & 1.00 & 0.98 & 0.98 & 0.96 \\
& Edge & 0.98 & 0.98 & 0.94 & 0.96 \\
\multirow{2}{*}{ Isolated C. } & Node & 0.46 & 0.47 & 0.63 & 0.67 \\
& Edge & 0.61 & 0.51 & 0.52 & 0.69 \\
\multirow{2}{*}{ Asymmetric } & Node & - & - & 0.59 & 0.62 \\
& Edge & - & - & 0.65 & 0.66 \\
\multirow{2}{*}{ AVG } & Node & 0.72 & 0.60 & 0.70 & 0.64 \\
& Edge & 0.76 & 0.58 & 0.65 & 0.61 \\
\bottomrule
\end{tabular}
}
    \label{shortest-path}
    \vspace{-0.2in}
\end{table}
% \begin{table*}[]
% \small
%     \centering
%     \begin{tabular}{cccccccccc}
% & & & 1-hop & 2-hop & 3-hop & 4-hop & k>5 & Asymmetric & AVG. scores \\
% ID names & Edge & acc & 1 & 1 & 0.9808 & 0.7805 & 0.6538 & 0.175 & 0.7166117391 \\
% & & f_acc & 0.64 & 0.6489 & 0.4808 & 0.12195 & 0 & 0.175 & 0.3810084783 \\
% & Node & acc & 1 & 0.9892 & 0.8868 & 0.6429 & 0.5167 & 0.3216 & 0.7369343478 \\
% & & f_acc & 0.7664 & 0.7849 & 0.4906 & 0.2143 & 0 & 0.3126 & 0.4981208696 \\
% Entity names & Edge & acc & 1 & 1 & 1 & 0.9524 & 0.9153 & 0.0754 & 0.7257565217 \\
% & & f_acc & 0.97 & 0.8085 & 0.5283 & 0.3333 & 0.0508 & 0.0754 & 0.4684993478 \\
% & Node & acc & 1 & 0.9681 & 0.9811 & 0.9048 & 0.8167 & 0.2374 & 0.7716641304 \\
% & & f_acc & 0.9907 & 0.8298 & 0.5472 & 0.2857 & 0.0333 & 0.2374 & 0.5415571739 \\
% \multirow{4}{*}{ Entity names+BFS cot } & Edge & acc & 1 & 0.9894 & 0.9811 & 0.9048 & 0.95 & 0.1717 & 0.763735 \\
% & & f_acc & 0.9907 & 0.8404 & 0.566 & 0.2619 & 0.0333 & 0.1717 & 0.5147454348 \\
% & Node & acc & 1 & 1 & 0.9245 & 0.8571 & 0.8333 & 0.4343 & 0.8521436957 \\
% & & f_acc & 0.9813 & 0.883 & 0.566 & 0.3333 & 0.0333 & 0.4343 & 0.6380371739 \\
% \end{tabular}
%     \caption{Caption}
%     \label{kg_results}
% \end{table*}

\begin{table*}[]
\vspace{-1cm}
\small
    \centering
        \caption{Entity connection on the knowledge graph}
        \setlength{\tabcolsep}{3pt}
\begin{tabular}{ccc|ccccccc}
\toprule
Node Naming & Des. & & 1-hop & 2-hop & 3-hop & 4-hop & k-hop, k>5 & Asymmetric & AVG. scores \\
\midrule
\multirow{4}{*}{ ID names } & \multirow{2}{*}{ Edge } & ACC & 1.0000 & 1.0000 & 0.9808 & 0.7805 & 0.6538 & 0.1750 & 0.7166 \\
& & $\mathrm{F_{acc}}$ & 0.6400 & 0.6489 & 0.4808 & 0.1220 & 0.0000 & 0.1750 & 0.3810 \\
& \multirow{2}{*}{ Node } & ACC & 1.0000 & 0.9892 & 0.8868 & 0.6429 & 0.5167 & 0.3216 & 0.7369 \\
& & $\mathrm{F_{acc}}$ & 0.7664 & 0.7849 & 0.4906 & 0.2143 & 0.0000 & 0.3126 & 0.4981 \\
\midrule
\multirow{4}{*}{ Entity names } & \multirow{2}{*}{ Edge } & ACC & 1.0000 & 1.0000 & 1.0000 & 0.9524 & 0.9153 & 0.0754 & 0.7258 \\
& & $\mathrm{F_{acc}}$ & 0.9700 & 0.8085 & 0.5283 & 0.3333 & 0.0508 & 0.0754 & 0.4685 \\
& \multirow{2}{*}{ Node } & ACC & 1.0000 & 0.9681 & 0.9811 & 0.9048 & 0.8167 & 0.2374 & 0.7717 \\
& & $\mathrm{F_{acc}}$ & 0.9907 & 0.8298 & 0.5472 & 0.2857 & 0.0333 & 0.2374 & 0.5416 \\
\midrule
\multirow{4}{*}{ \parbox{1.7cm}{Entity names + BFS COT} } & \multirow{2}{*}{ Edge } & ACC & 1.0000 & 0.9894 & 0.9811 & 0.9048 & 0.9500 & 0.1717 & 0.7637 \\
& & $\mathrm{F_{acc}}$ & 0.9907 & 0.8404 & 0.5660 & 0.2619 & 0.0333 & 0.1717 & 0.5147 \\
& \multirow{2}{*}{ Node } & ACC & 1.0000 & 1.0000 & 0.9245 & 0.8571 & 0.8333 & 0.4343 & 0.8521 \\
& & $\mathrm{F_{acc}}$ & 0.9813 & 0.8830 & 0.5660 & 0.3333 & 0.0333 & 0.4343 & 0.6380 \\
\bottomrule
\end{tabular}
    \label{kg_results}
    \vspace{-0.2in}
\end{table*}
\vspace{-0.15cm}
\subsection{The shortest-path problem}
\label{shortest-path_sec}
The shortest-path problem is another essential task theoretically proven to be achievable by LLMs, yet it fails in practice. Compared to the graph connectivity task, it is more challenging because it requires not only determining whether nodes are connected but also calculating edge weights to identify the shortest path among multiple potential solutions.  Next, we explore if the varied performance of LLMs across different connectivity types is also applicable to the shortest-path problem. The details of experiment settings are in Appendix~\ref{sec:shortest-path-experiment_setting}.

% \paragraph{Experimental setup}
% We study the shortest-path problem using the Easy datasets from the unweighted graphs as mentioned in Section~\ref{connectivity}. For the weighted graphs, we applied similar strategies that were used in undirected graph generations to generate the directed and undirected graph datasets.
% The directed graph datasets include two types, whether there are negative edges in the graphs. Appendix~\ref{graph_connect_datasets} Table~\ref{all_dataset} shows the details. The graph structure descriptions are shown in Appendix~\ref{description_method}

We use GPT-4 to illustrate an example of the shortest-path problem. Table~\ref{shortest-path} displays the results of LLMs' performance. The findings for the shortest path problem align with our observations from graph connectivity, where performance diminishes as the path length (k-hop) increases. Moreover, undirected graphs consistently outperform directed graphs. We observe a significant difference in LLM performance between datasets with weighted edges and those without. This suggests that LLMs might overlook or misrepresent edge weights in the text.

\subsection{Entity connection on the knowledge graph}
\label{kg_sec}
%\jt{please add some motivation examples why connectivity in KGs are important}
To determine whether our findings can apply to real-world applications, we performed the entity connection on the knowledge graph using WN18RR~\cite{shang2019end} dataset. The details of dataset construction are provided in Appendix~\ref{description_method}.

% \paragraph{Dataset} We used WN18RR \cite{shang2019end} as the base dataset, which provides both ID names and Entity names. The ID names consist of strings of random numbers, and Entity names are used as specific and meaningful identifiers. From its training set, we randomly selected 150 subgraphs based on ego graphs with a depth of 3. Within each subgraph, we identified two nodes with the longest paths and segmented the paths into $k'$-hops. This strategy allowed us to generate $k'$ question-answer pairs, ranging from 1-hop to $k'$-hop.
% Table~\ref{app_datasets} shows the details of our sampled dataset. We take both Node List and Edge List in the experiment. The description examples can be found in Appendix~\ref{description_method}.

We use GPT-4 to evaluate the connections, and summarize the results in Table~\ref{kg_results}. 
The performance trends align with Section~\ref{sec:connect_results} and Section~\ref{shortest-path_sec}. Specifically, the performance declines with $K$ increasing in the $K$-hop setting, and Node List descriptions outperform Edge List descriptions. Notably, LLMs demonstrate improved performance with meaningful node naming. Furthermore, incorporating BFS into the prompt results in significant improvements. Detailed analyses are provided in Appendix~\ref{naming} and Appendix~\ref{strategies}.

\section{Other factors for graph reasoning}
\label{sec:reasons}

\subsection{Impact of training data and model scale}

% \jt{you use "training GPT for graph task", should we use fine-tune, meanwhile, the data should be fine-tuning data?}

% \jt{you mention this new contribution in the intro but not in the abstract, please update the abs too}

To explore the impact of training data and model scale on LLMs' graph reasoning, we train GPT-2 from scratch to perform $k$-hop reasoning. Specifically, we use the Medium (M), Small (S), and Baby (B) versions, as statistics are outlined in Table~\ref{tab:model_scale1}. 
We create a new dataset different from the previous sections, focusing on $K$-hop connections within directed graphs consisting of 5 to 15 nodes. The training dataset contains 210,000 unique question-answer pairs, with 20,000 reserved for validation and 10,000 for testing. 

We focus on 3-hop and 5-hop connections, requiring GPT-2 to solve the shortest path problem using simplified graph descriptions. An example is shown in Table~\ref{tab:des}. Additional training details are provided in Appendix~\ref{sec:training_details}.

\begin{figure}[ht!]
% \vspace{-0.4cm}
    \centering
    \begin{minipage}[b]{0.23\textwidth}
    \includegraphics[width=0.9\linewidth]{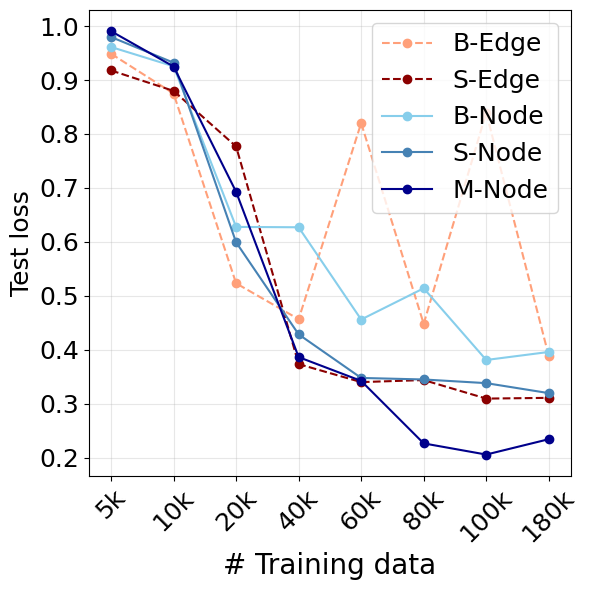}
    \caption{3-hop results}
    \label{fig:hop3_scale}
    \end{minipage}
    \begin{minipage}[b]{0.23\textwidth}
    \includegraphics[width=0.9\linewidth]{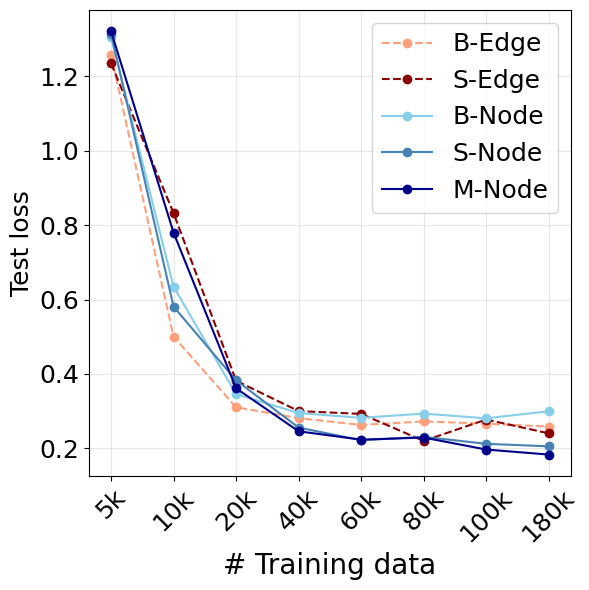}
    \caption{5-hop results}
    \label{fig:hop5_scale}
    \end{minipage}
\end{figure}

To analyze the effect of the training data scale, we vary the amount of training data from 1,000 to 180,000 and report the test loss on the test set. Besides, we also evaluate various versions of GPT-2 and different graph descriptions. The results are shown in Figure~\ref{fig:hop3_scale} and Figure~\ref{fig:hop5_scale}.

We observe that the reasoning ability of LLMs is strongly correlated with both the scale of the training data and the models, regardless of using Node List or Edge List. The test loss decreases significantly for both 3-hop and 5-hop reasoning tasks, demonstrating that GPT-2 architectures can perform reasoning tasks with sufficient model size and training data. The final test accuracy achieved about 80\%. Moreover, Node List is more stable than Edge List, likely because the Edge List contains longer sentences (up to 400 characters) compared to the Node List (up to 200 characters). %\jt{do you have reference for the following statement?} 
This increased length may hinder the transformer's ability to process the long context~\cite{long_context_transformer}.

Additionally, we evaluate the impact of data on fine-tuning. We use Llama3.2-3B as the backbone model, apply LoRA for fine-tuning to enhance shortest-path reasoning, and demonstrate a similar effect with increasing the training data. The whole details are shown in Appendix~\ref{sec:finetune}. 

\subsection{Different reasoning processes in Node List and Edge List}
\begin{figure}[htb!]
    \centering
    \includegraphics[width=0.8\linewidth]{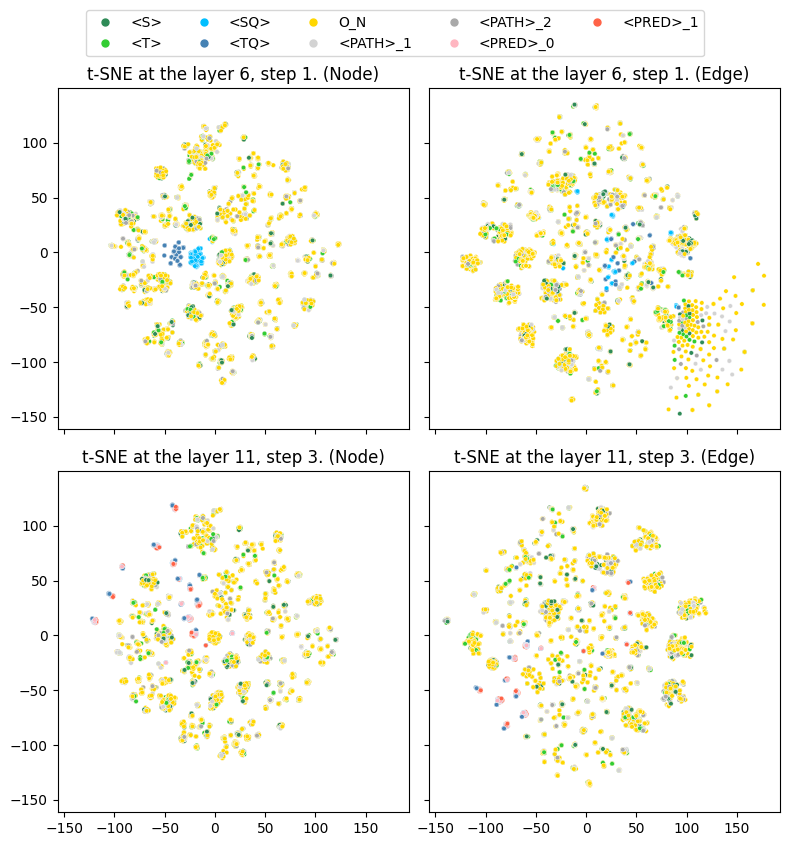}
    \caption{t-SNE results on the different layers and steps}
    \label{fig:vis_small}
\end{figure}

%The performance of LLMs differs significantly between Node and Edge Lists.
To gain deeper insights into how the LLM performs reasoning on a graph, we drew inspiration from syntax analysis in language models~\cite{bert_vis} to the reasoning tasks. Specifically, we applied t-SNE clustering to the outputs of various attention layers in a small-scale GPT-2 model at different reasoning steps. We define special labels for the nodes in the sentence. The details are illustrated in Appendix~\ref{sec:app_reasoning_process}. %In the graph description, \textless{}S\textgreater\ and \textless{}T\textgreater\ are designated as the source and target nodes, highlighted in green. \textless{}PATH\textgreater$_i$ in gray, denotes the $i$-th node on the path from the source node to the target node, while \textless{}O\_N\textgreater\ represents other nodes, shown in yellow. In the question, the source and target nodes are labeled as \textless{}SQ\textgreater\ and \textless{}TQ\textgreater\ in blue, respectively. When the models predict the $i+1$-th nodes, they require previous information, denoted as \textless{}PRED$_i$\textgreater\ in red. 
To analyze the reasoning process, we selected 200 correct samples from the test set. The results are presented in Figure~\ref{fig:vis_small}. The Appendix Figure~\ref{fig:tsne_adj} and Figure~\ref{fig:tsne_edge} provides more comprehensive results.

% \begin{figure*}
%     \centering
% \vspace{-1cm}

%     \includegraphics[width=0.9\linewidth]{latex/figs/tsne_adj.png}
%     \caption{The t-SNE results on the Node List.}
%     \label{fig:tsne_adj}
%     \vspace{-0.4cm}
% \end{figure*}

% \begin{figure*}
%     \centering
%     \includegraphics[width=0.9\linewidth]{latex/figs/tsne_edge.png}
%     \caption{The t-SNE results on the Edge List.}
%     \label{fig:tsne_edge}
% \vspace{-0.5cm}

% \end{figure*}

The observations can be summarized as follows:
(1) GPT-2 learns the node combination patterns in both Node and Edge Lists. In the final layer, the model reorganizes these combinations to identify the source (light blue), target (deep blue), and predicted nodes (red). The observed pattern combination phenomenon suggests that, rather than directly extracting path information from the given sentences, the GPT model relies on learned relationships among different nodes. This reliance also explains the accuracy drop in the connectivity task when node IDs are replaced with random numbers or characters, as such disruptions interfere with the model's learned combinations. 
(2) GPT-2 exhibits different reasoning processes when using Node List and Edge List as graph descriptions. With the Node List, GPT-2 easily captures the source and target nodes in the middle layers, subsequently identifying path patterns based on the source and target in the question. In contrast, when using the Edge List, the model tends to skip this intermediate step, instead directly matching the source and target nodes to its learned node combination patterns.

% \subsection{Influence of language tokens}
\section{Conclusion}

In this paper, we focus on the graph reasoning ability of LLMs. Recently, there exists a discrepancy between theoretical potential and poor empirical performance.
To bridge this gap, we construct a balanced and comprehensive benchmarking, and conclude that graph reasoning ability is influenced by various node connectivity types, graph sizes, graph descriptions, and node naming methods. Moreover, we also demonstrate the impacts of training data, model size and fine-tuning on graph reasoning ability. 
These findings offer valuable insights to enhance LLMs in graph reasoning tasks.

\noindent  \textbf{Limitation:}
Our computational resources are limited, which poses challenges in fully exploring the upper bounds of training or fine-tuning LLMs for graph reasoning tasks.
% Currently we explore the graph reasoning capabilities of LLMs without fine-tuning. Our future research will focus on incorporating effective fine-tuning strategies and novel approaches to  enhance the graph reasoning ability.

\bibliography{custom}

\appendix

\label{sec:appendix}

\newpage

% \begin{table*}[]
%     \centering
%     \begin{tabular}{cccccccccccccc}
% & & & & 1-hop & 2-hop & 3-hop & 4-hop & 5-hop & 6-hop & isolated node graphs & same component (noniso) & diff component & others \\
% undirected & GPT-4 & tiny(5-7) & incv2 & 1 & 1 & 1 & 0.833 & 1 & - & - & 0.6 & & - \\
% & & & adj & 1 & 1 & 1 & 0.833 & 1 & - & - & 0.6 & & - \\
% & & small(8-15) & incv2 & 1 & 1 & 1 & 1 & 1 & 1 & - & 0.353 & & 1 \\
% & & & adj & 1 & 1 & 1 & 1 & 0.667 & 1 & - & 0.412 & & 1 \\
% & & medium(16-25) & incv2 & 1 & 1 & 1 & 1 & 1 & 1 & - & 0.115 & & 1 \\
% & & & adj & 1 & 1 & 1 & 1 & 1 & 1 & - & 0.712 & & 0.667 \\
% & & large(26-35) & incv2 & 1 & 1 & 1 & 1 & 1 & - & - & 0.04 & & 0.75 \\
% & & & adj & 1 & 1 & 1 & 1 & 1 & - & - & 0.36 & & 0.5 \\
% & GPT-3 & tiny & incv2 & 1 & 0.884 & 0.364 & 0 & 0 & - & - & 0 & & - \\
% & & & adj & 1 & 0.791 & 0.182 & 0.333 & 0 & - & - & 0.4 & & - \\
% & & small & incv2 & 0.976 & 0.911 & 0.917 & 0.636 & 0.667 & 1 & - & 0.559 & & 1 \\
% & & & adj & 1 & 0.911 & 0.75 & 0.455 & 0.333 & 1 & - & 0.5 & & 0 \\
% & & medium & incv2 & 1 & 0.98 & 0.842 & 0.667 & 1 & 0 & - & 0.673 & & 0 \\
% & & & adj & 0.973 & 0.96 & 0.632 & 1 & 1 & 0.5 & - & 0.423 & & 0 \\
% & & large & incv2 & 1 & 0.98 & 0.85 & 0.8 & 0.5 & - & - & 0.36 & & 1 \\
% & & & adj & 1 & 0.959 & 0.9 & 0.6 & 1 & - & - & 0.3 & & 0.75 \\
% \end{tabular}
%     \caption{Caption}
%     \label{baseline_all}
% \end{table*}

\begin{table*}[]
    \centering
    \tiny
    \setlength{\tabcolsep}{8pt}
\begin{tabular}{c@{\hskip 0in}c@{\hskip 0in}c@{\hskip 0in}c|cccccccccc}
\toprule
Directivity & Model & Difficulty & Des. & 1-hop & 2-hop & 3-hop & 4-hop & 5-hop & 6-hop & Singleton & Isolated C. & Asymmetric & k-hop k>6 \\
\midrule
\multicolumn{4}{c|}{Dataset} & \multicolumn{10}{c}{GraphInstruct} \\
\midrule
\multirow{16}{*}{ Undirected } & \multirow{8}{*}{ GPT-4 } & \multirow{2}{*}{ Tiny } & Node & 1.00 & 1.00 & 1.00 & 0.83 & 1.00 & - & - & 0.60 & - & - \\
& & & Edge & 1.00 & 1.00 & 1.00 & 0.83 & 1.00 & - & - & 0.60 & - & - \\
& & \multirow{2}{*}{ Easy } & Node & 1.00 & 1.00 & 1.00 & 1.00 & 1.00 & 1.00 & - & 0.35 & - & 1.00 \\
& & & Edge & 1.00 & 1.00 & 1.00 & 1.00 & 0.67 & 1.00 & - & 0.41 & - & 1.00 \\
& & \multirow{2}{*}{ Med } & Node & 1.00 & 1.00 & 1.00 & 1.00 & 1.00 & 1.00 & - & 0.12 & - & 1.00 \\
& & & Edge & 1.00 & 1.00 & 1.00 & 1.00 & 1.00 & 1.00 & - & 0.71 & - & 0.67 \\
& & \multirow{2}{*}{ Hard } & Node & 1.00 & 1.00 & 1.00 & 1.00 & 1.00 & - & - & 0.04 & - & 0.75 \\
& & & Edge & 1.00 & 1.00 & 1.00 & 1.00 & 1.00 & - & - & 0.36 & - & 0.50 \\
& \multirow{8}{*}{ GPT-3 } & \multirow{2}{*}{ Tiny } & Node & 1.00 & 0.88 & 0.36 & 0.00 & 0.00 & - & - & 0.00 & - & - \\
& & & Edge & 1.00 & 0.79 & 0.18 & 0.33 & 0.00 & - & - & 0.40 & - & - \\
& & \multirow{2}{*}{ Easy } & Node & 0.98 & 0.91 & 0.92 & 0.64 & 0.67 & 1.00 & - & 0.56 & - & 1.00 \\
& & & Edge & 1.00 & 0.91 & 0.75 & 0.45 & 0.33 & 1.00 & - & 0.50 & - & 0.00 \\
& & \multirow{2}{*}{ Med } & Node & 1.00 & 0.98 & 0.84 & 0.67 & 1.00 & 0.00 & - & 0.67 & - & 0.00 \\
& & & Edge & 0.97 & 0.96 & 0.63 & 1.00 & 1.00 & 0.50 & - & 0.42 & - & 0.00 \\
& & \multirow{2}{*}{ Hard } & Node & 1.00 & 0.98 & 0.85 & 0.80 & 0.50 & - & - & 0.36 & - & 1.00 \\
& & & Edge & 1.00 & 0.96 & 0.90 & 0.60 & 1.00 & - & - & 0.30 & - & 0.75 \\
\midrule
\multirow{16}{*}{ Directed } & \multirow{8}{*}{ GPT-4 } & \multirow{2}{*}{ Tiny } & Node & 1.00 & 0.92 & 0.14 & - & - & - & - & 1.00 & 0.95 & - \\
& & & Edge & 1.00 & 0.85 & 0.43 & - & - & - & - & 1.00 & 0.97 & - \\
& & \multirow{2}{*}{ Easy } & Node & 1.00 & 0.93 & 1.00 & 0.67 & - & - & - & - & 0.91 & - \\
& & & Edge & 1.00 & 0.64 & 0.83 & 0.33 & - & - & - & - & 0.91 & - \\
& & \multirow{2}{*}{ Med } & Node & 0.78 & 0.71 & 0.60 & 1.00 & 1.00 & - & - & - & 0.82 & - \\
& & & Edge & 0.89 & 0.71 & 1.00 & 0.50 & 1.00 & - & - & - & 0.78 & - \\
& & \multirow{2}{*}{ Hard } & Node & 0.90 & 0.88 & 0.60 & 1.00 & 1.00 & 1.00 & - & - & 0.77 & - \\
& & & Edge & 1.00 & 0.88 & 0.60 & 1.00 & 1.00 & 1.00 & - & - & 0.83 & - \\
& \multirow{8}{*}{ GPT-3 } & \multirow{2}{*}{ Tiny } & Node & 0.94 & 0.92 & 1.00 & - & - & - & - & 1.00 & 0.26 & - \\
& & & Edge & 1.00 & 1.00 & 0.71 & - & - & - & - & 1.00 & 0.27 & - \\
& & \multirow{2}{*}{ Easy } & Node & 0.77 & 0.93 & 0.83 & 1.00 & - & - & - & - & 0.19 & - \\
& & & Edge & 1.00 & 0.93 & 0.83 & 1.00 & - & - & - & - & 0.31 & - \\
& & \multirow{2}{*}{ Med } & Node & 1.00 & 1.00 & 1.00 & 0.50 & 1.00 & - & - & - & 0.33 & - \\
& & & Edge & 1.00 & 0.79 & 0.80 & 1.00 & 1.00 & - & - & - & 0.42 & - \\
& & \multirow{2}{*}{ Hard } & Node & 1.00 & 0.88 & 1.00 & 0.00 & 1.00 & 1.00 & - & - & 0.22 & - \\
& & & Edge & 1.00 & 0.88 & 0.90 & 0.00 & 0.50 & 1.00 & - & - & 0.37 & - \\
\bottomrule
\end{tabular}

\caption{Baseline result of zero-shot accuracy on GraphInstruct dataset.}
\label{baseline_all}

\setlength{\tabcolsep}{5pt}
\begin{tabular}{c@{\hskip 0in}c@{\hskip 0in}c|ccccccccccccc}
\toprule
\multirow{2}{*}{ Difficulty } & \multirow{2}{*}{ Model } & \multirow{2}{*}{ Des. } & \multicolumn{3}{c}{k-hop, 1$\leq$k$\leq$2} & \multicolumn{3}{c}{k-hop, 3$\leq$k$\leq$4} & \multicolumn{3}{c}{5-hop} & Singleton & Isolated C. & \multirow{2}{*}{ AVG. ACC } & \multirow{2}{*}{ AVG. $\mathrm{F_{acc}}$ } \\
& & & ACC & $\mathrm{F_{acc}}$ & PCR & ACC & $\mathrm{F_{acc}}$ & PCR & ACC & $\mathrm{F_{acc}}$ & PCR & $\mathrm{F_{acc}}$ & $\mathrm{F_{acc}}$ & & \\
\midrule
\multirow{4}{*}{ Tiny } & \multirow{2}{*}{ GPT-3 } & Node & 1.00 & 0.99 & 0.90 & 0.82 & 0.82 & 0.95 & 1.00 & 1.00 & 1.00 & - & 0.00 & 0.93 & 0.92 \\
& & Edge & 1.00 & 0.96 & 0.87 & 0.65 & 0.65 & 0.96 & 1.00 & 1.00 & 1.00 & - & 0.00 & 0.91 & 0.87 \\
& \multirow{2}{*}{ GPT-4 } & Node & 1.00 & 1.00 & 1.00 & 1.00 & 1.00 & 1.00 & 1.00 & 1.00 & 1.00 & - & 0.80 & 0.99 & 0.99 \\
& & Edge & 1.00 & 0.98 & 0.99 & 1.00 & 1.00 & 1.00 & 1.00 & 1.00 & 1.00 & - & 0.80 & 0.99 & 0.97 \\
\midrule
\multirow{4}{*}{ Easy } & \multirow{2}{*}{ GPT-3 } & Node & 1.00 & 0.97 & 0.85 & 1.00 & 0.91 & 0.95 & 1.00 & 1.00 & 0.94 & - & 0.09 & 0.79 & 0.75 \\
& & Edge & 1.00 & 0.93 & 0.80 & 0.87 & 0.61 & 0.82 & 1.00 & 0.67 & 0.77 & - & 0.00 & 0.75 & 0.66 \\
& \multirow{2}{*}{ GPT-4 } & Node & 1.00 & 0.97 & 0.99 & 1.00 & 1.00 & 1.00 & 1.00 & 1.00 & 1.00 & - & 0.68 & 0.92 & 0.90 \\
& & Edge & 1.00 & 0.98 & 0.99 & 1.00 & 1.00 & 0.99 & 1.00 & 1.00 & 0.94 & - & 0.74 & 0.94 & 0.92 \\
\midrule
\multirow{4}{*}{ Medium } & \multirow{2}{*}{ GPT-3 } & Node & 0.99 & 0.98 & 0.69 & 1.00 & 1.00 & 0.90 & 1.00 & 0.00 & 0.00 & - & 0.00 & 0.68 & 0.66 \\
& & Edge & 0.99 & 0.86 & 0.72 & 0.96 & 0.68 & 0.88 & 1.00 & 0.50 & 1.00 & - & 0.02 & 0.68 & 0.57 \\
& \multirow{2}{*}{ GPT-4 } & Node & 1.00 & 0.92 & 0.98 & 1.00 & 0.80 & 0.99 & 1.00 & 1.00 & 1.00 & - & 0.56 & 0.86 & 0.79 \\
& & Edge & 1.00 & 0.98 & 0.96 & 1.00 & 0.88 & 0.98 & 1.00 & 1.00 & 1.00 & - & 0.77 & 0.93 & 0.90 \\
\midrule
\multirow{4}{*}{ Hard } & \multirow{2}{*}{ GPT-3 } & Node & 1.00 & 0.94 & 0.63 & 1.00 & 0.76 & 0.85 & 1.00 & 0.00 & 0.00 & - & 0.10 & 0.71 & 0.63 \\
& & Edge & 1.00 & 0.78 & 0.56 & 1.00 & 0.56 & 0.81 & 1.00 & 0.00 & 0.00 & - & 0.08 & 0.70 & 0.51 \\
& \multirow{2}{*}{ GPT-4 } & Node & 1.00 & 0.87 & 0.94 & 1.00 & 1.00 & 0.93 & 1.00 & 0.50 & 0.71 & - & 0.34 & 0.79 & 0.72 \\
& & Edge & 1.00 & 0.87 & 0.90 & 1.00 & 0.96 & 0.93 & 1.00 & 1.00 & 1.00 & - & 0.62 & 0.88 & 0.81 \\
\bottomrule
\end{tabular}

\caption{Undirected Baseline result of ACC and $\mathrm{F_{acc}}$.}
\label{baseline_all_pathundirected}

\setlength{\tabcolsep}{4pt}
\begin{tabular}{c@{\hskip 0in}c@{\hskip 0in}c|cccccccccccccc}
\toprule
\multirow{2}{*}{ Difficulty } & \multirow{2}{*}{ Model } & \multirow{2}{*}{ Des. } & \multicolumn{3}{c}{k-hop, 1$\leq$k$\leq$2} & \multicolumn{3}{c}{k-hop, 3$\leq$k$\leq$4} & \multicolumn{3}{c}{5-hop} & Singleton & Isolated C. & Asymmetric & \multirow{2}{*}{ AVG. ACC } & \multirow{2}{*}{ AVG. $\mathrm{F_{acc}}$ } \\
& & & ACC & $\mathrm{F_{acc}}$ & PCR & ACC & $\mathrm{F_{acc}}$ & PCR & ACC & $\mathrm{F_{acc}}$ & PCR & $\mathrm{F_{acc}}$ & $\mathrm{F_{acc}}$ & $\mathrm{F_{acc}}$ & & \\
\midrule
\multirow{4}{*}{ Tiny } & \multirow{2}{*}{ GPT-3 } & Node & 1.00 & 0.05 & 0.62 & 1.00 & 0.00 & 0.00 & - & - & - & - & - & 0.06 & 0.20 & 0.06 \\
& & Edge & 1.00 & 0.94 & 0.99 & 1.00 & 0.71 & 0.95 & - & - & - & - & 1.00 & 0.04 & 0.25 & 0.22 \\
& \multirow{2}{*}{ GPT-4 } & Node & 1.00 & 1.00 & 0.96 & 1.00 & 0.86 & 1.00 & - & - & - & - & 1.00 & 0.88 & 0.90 & 0.90 \\
& & Edge & 1.00 & 1.00 & 1.00 & 1.00 & 1.00 & 1.00 & - & - & - & - & 1.00 & 0.85 & 0.88 & 0.88 \\
\midrule
\multirow{4}{*}{ Easy } & \multirow{2}{*}{ GPT-3 } & Node & 1.00 & 0.00 & 0.00 & 1.00 & 0.00 & 0.00 & - & - & - & - & - & 0.04 & 0.17 & 0.03 \\
& & Edge & 0.96 & 0.89 & 0.93 & 1.00 & 0.78 & 0.83 & - & - & - & - & - & 0.07 & 0.28 & 0.26 \\
& \multirow{2}{*}{ GPT-4 } & Node & 1.00 & 0.96 & 1.00 & 1.00 & 0.89 & 1.00 & - & - & - & - & - & 0.87 & 0.90 & 0.89 \\
& & Edge & 1.00 & 1.00 & 0.95 & 1.00 & 0.67 & 0.95 & - & - & - & - & - & 0.81 & 0.86 & 0.84 \\
\midrule
\multirow{4}{*}{ Medium } & \multirow{2}{*}{ GPT-3 } & Node & 1.00 & 0.00 & 0.00 & - & - & - & - & - & - & - & - & 0.08 & 0.19 & 0.07 \\
& & Edge & 1.00 & 0.70 & 0.88 & 1.00 & 0.29 & 0.51 & 1.00 & 0.00 & 0.00 & - & - & 0.10 & 0.32 & 0.22 \\
& \multirow{2}{*}{ GPT-4 } & Node & 0.96 & 0.83 & 0.87 & 1.00 & 0.43 & 1.00 & 1.00 & 0.00 & 0.00 & - & - & 0.67 & 0.74 & 0.68 \\
& & Edge & 1.00 & 0.87 & 0.91 & 1.00 & 0.57 & 0.97 & 1.00 & 0.00 & 0.00 & - & - & 0.67 & 0.75 & 0.70 \\
\midrule
\multirow{4}{*}{ Hard } & \multirow{2}{*}{ GPT-3 } & Node & 1.00 & 0.00 & 0.00 & 1.00 & 0.00 & 0.00 & - & - & - & - & - & 0.17 & 0.32 & 0.14 \\
& & Edge & 1.00 & 0.70 & 0.74 & 1.00 & 0.36 & 0.85 & 1.00 & 0.50 & 1.00 & - & - & 0.12 & 0.37 & 0.26 \\
& \multirow{2}{*}{ GPT-4 } & Node & 1.00 & 0.74 & 0.93 & 0.91 & 0.45 & 0.82 & 1.00 & 0.00 & 0.00 & - & - & 0.59 & 0.70 & 0.60 \\
& & Edge & 1.00 & 0.81 & 0.95 & 1.00 & 0.73 & 0.87 & 1.00 & 0.50 & 0.83 & - & - & 0.67 & 0.76 & 0.70 \\
\bottomrule
\end{tabular}

\caption{Directed Baseline result of ACC and $\mathrm{F_{acc}}$. '-' indicates no data.}

\label{baseline_all_pathdirected}
\end{table*}

\begin{table}[]
    \centering
    \tiny
\begin{tabular}{c@{\hskip 0in}c@{\hskip 0in}c|cccc}
\toprule
& Subdataset & Des. & 0-shot & few-shot & 0-dijkstra & cot-dijkstra \\
\midrule
\multirow{12}{*}{ unweighted } & \multirow{2}{*}{ 1$\leq$k$\leq$2 hops } & Node & 0.88 & 0.91 & 0.92 & 0.96 \\
& & Edge & 0.89 & 0.82 & 0.87 & 0.96 \\
& \multirow{2}{*}{ 3$\leq$k$\leq$4 hops } & Node & 0.87 & 0.90 & 0.87 & 0.94 \\
& & Edge & 0.81 & 0.86 & 0.83 & 0.85 \\
& \multirow{2}{*}{ 5-hops } & Node & 0.88 & 0.78 & 0.78 & 0.86 \\
& & Edge & 0.76 & 0.68 & 0.74 & 0.82 \\
& \multirow{2}{*}{ Singleton } & Node & 1.00 & 1.00 & 0.86 & 1.00 \\
& & Edge & 0.98 & 1.00 & 0.84 & 0.96 \\
& \multirow{2}{*}{ I.C. } & Node & 0.46 & 0.52 & 0.58 & 0.70 \\
& & Edge & 0.61 & 0.37 & 0.64 & 0.74 \\
& \multirow{2}{*}{ AVG } & Node & 0.72 & 0.75 & 0.75 & 0.84 \\
& & Edge & 0.76 & 0.65 & 0.75 & 0.81 \\
\midrule
\multirow{12}{*}{ Weighted } & \multirow{2}{*}{ 1$\leq$k$\leq$2 hops } & Node & 0.80 & 0.75 & 0.75 & 0.81 \\
& & Edge & 0.70 & 0.66 & 0.65 & 0.73 \\
& \multirow{2}{*}{ 3$\leq$k$\leq$4 hops } & Node & 0.52 & 0.58 & 0.59 & 0.65 \\
& & Edge & 0.47 & 0.47 & 0.48 & 0.64 \\
& \multirow{2}{*}{ 5-hops } & Node & 0.54 & 0.48 & 0.54 & 0.58 \\
& & Edge & 0.44 & 0.52 & 0.44 & 0.50 \\
& \multirow{2}{*}{ Singleton } & Node & 0.98 & 0.92 & 0.80 & 0.84 \\
& & Edge & 0.98 & 1.00 & 0.76 & 0.98 \\
& \multirow{2}{*}{ I.C. } & Node & 0.47 & 0.39 & 0.35 & 0.53 \\
& & Edge & 0.51 & 0.32 & 0.46 & 0.57 \\
& \multirow{2}{*}{ AVG } & Node & 0.60 & 0.56 & 0.54 & 0.65 \\
& & Edge & 0.58 & 0.51 & 0.53 & 0.65 \\
\bottomrule
\end{tabular}
    \caption{Shortest path result with strategy}
    \label{strategy2}
\end{table}
\begin{table}[]
    \centering
    \small
    \setlength{\tabcolsep}{3pt}
\begin{tabular}{c|ccc}
\hline
Connectivity types & \# Sample & AVG. \# Node & AVG. \# Edge \\
\hline
1-hop & 107 & 82 & 199 \\
2-hop & 64 & 104 & 257 \\
3-hop & 53 & 139 & 347 \\
4-hop & 42 & 145 & 363 \\
k-hop (k$\geq$5) & 60 & 201 & 521 \\
Asymmetric & 198 & 49 & 106 \\
\hline
\end{tabular}
    \caption{Knowledge graph dataset.}
    \label{app_datasets}
\end{table}

\begin{table*}[]
\tiny
\setlength{\tabcolsep}{4pt}
\begin{tabular}{cc|cccccccc}
\toprule
& & 1-hop & 2-hop & 3-hop & 4-hop & 5-hop & Singleton & Isolated C. & Asymmetric \\
\midrule
& Dataset & \multicolumn{8}{c}{GraphInstruct Dataset (Tiny/Easy/Med/Hard)} \\
\multirow{3}{*}{ Undirected } & \# Sample & 51 / 41 / 37 / 29 & 43 / 45 / 50 / 49 & 11 / 12 / 19 / 20 & 6 / 11 / 6 / 5 & 1 / 3 / 2 / 2 & - / - / - / - & 5 / 34 / 52 / 50 & - \\
& AVG. \# Node & 6 / 12 / 21 / 30 & 6 / 12 / 21 / 30 & 6 / 12 / 20 / 31 & 6 / 11 / 22 / 31 & 7 / 14 / 20 / 28 & - / - / - / - & 7 / 12 / 20 / 31 & - \\
& AVG. \# Edge & 8 / 24 / 77 / 181 & 7 / 22 / 68 / 125 & 5 / 14 / 37 / 62 & 6 / 10 / 33 / 49 & 7 / 14 / 20 / 28 & - / - / - / - & 7 / 12 / 20 / 31 & - \\
\multirow{3}{*}{ Directed } & \# Sample & 18 / 13 / 9 / 10 & 13 / 14 / 14 / 17 & 7 / 6 / 5 / 10 & - / 3 / 2 / 1 & - / - / 1 / 2 & - / - / - / - & 1 / - / - / - & 144 / 116 / 98 / 100 \\
& AVG. \# Node & 6 / 12 / 21 / 30 & 6 / 12 / 21 / 31 & 7 / 12 / 20 / 29 & - / 15 / 20 / 31 & - / - / 19 / 32 & - / - / - / - & 6 / - / - / - & 6 / 11 / 21 / 31 \\
& AVG. \# Edge & 15 / 44 / 117 / 194 & 15 / 38 / 123 / 220 & 14 / 30 / 80 / 140 & - / 28 / 50 / 56 & - / - / 47 / 70 & - / - / - / - & 4 / - / - / - & 10 / 24 / 45 / 73 \\
\midrule
& Dataset & \multicolumn{8}{c}{NLGraph Dataset (Easy/Med/Hard)} \\
\multirow{3}{*}{ Undirected } & \# Sample & 137 / 417 / 163 & 36 / 146 / 152 & 3 / 30 / 21 & - / 5 / 4 & - / 2 / - & 51 / 106 / 42 & 125 / 494 / 298 & - \\
& AVG. \# Node & 7 / 19 / 31 & 8 / 19 / 31 & 9 / 19 / 30 & - / 17 / 32 & - / 20 / - & 7 / 17 / 31 & 7 / 19 / 31 & - \\
& AVG. \# Edge & 11 / 78 / 138 & 8 / 47 / 103 & 7 / 26 / 56 & - / 24 / 44 & - / 20 / - & 7 / 49 / 127 & 11 / 71 / 103 & - \\
\midrule
& Dataset & \multicolumn{8}{c}{Our Dataset with Unweighted Edge Graphs (Easy/Med/Hard)} \\
\multirow{3}{*}{ Undirected } & \# Sample & 50 / 50 / 50 & 50 / 50 / 50 & 50 / 50 / 50 & 50 / 50 / 50 & 50 / 50 / 50 & 50 / 50 / 50 & 200 / 200 / 200 & - \\
& AVG. \# Node & 10 / 21 / 30 & 10 / 21 / 31 & 11 / 21 / 30 & 11 / 20 / 31 & 11 / 20 / 30 & 11 / 20 / 31 & 11 / 21 / 31 & - \\
& AVG. \# Edge & 32 / 104 / 229 & 33 / 112 / 215 & 26 / 83 / 158 & 21 / 51 / 146 & 17 / 43 / 90 & 35 / 93 / 198 & 20 / 60 / 113 & - \\
\multirow{3}{*}{ Directed } & \# Sample & 50 / 50 / 50 & 50 / 50 / 50 & 50 / 50 / 50 & 50 / 50 / 50 & 50 / 50 / 50 & 50 / 50 / 50 & 100 / 100 / 100 & 100 / 100 / 100 \\
& AVG. \# Node & 10 / 20 / 30 & 10 / 20 / 31 & 10 / 20 / 31 & 10 / 20 / 31 & 11 / 20 / 30 & 10 / 21 / 31 & 11 / 21 / 31 & 11 / 21 / 31 \\
& AVG. \# Edge & 64 / 191 / 514 & 57 / 191 / 479 & 49 / 170 / 409 & 38 / 131 / 251 & 32 / 89 / 185 & 45 / 162 / 466 & 35 / 102 / 188 & 57 / 120 / 279 \\
\midrule
& Dataset & \multicolumn{8}{c}{Our Dataset Wtih Positive Weighted Edge Graphs (Easy/Med/Hard)} \\
\multirow{3}{*}{ Undirected } & \# Sample & 50 / 50 / 50 & 50 / 50 / 50 & 50 / 50 / 50 & 50 / 50 / 50 & 50 / 50 / 50 & 50 / 50 / 50 & 200 / 200 / 200 & - \\
& AVG. \# Node & 10 / 20 / 30 & 10 / 20 / 30 & 10 / 20 / 30 & 11 / 20 / 30 & 11 / 20 / 30 & 11 / 20 / 30 & 11 / 20 / 30 & - \\
& AVG. \# Edge & 32 / 104 / 231 & 33 / 109 / 221 & 26 / 81 / 192 & 21 / 54 / 119 & 19 / 42 / 96 & 30 / 91 / 233 & 18 / 56 / 115 & - \\
\multirow{3}{*}{ Directed } & \# Sample & 50 / 50 / 50 & 50 / 50 / 50 & 50 / 50 / 50 & 50 / 50 / 50 & 50 / 50 / 50 & 50 / 50 / 50 & 100 / 100 / 100 & 100 / 100 / 100 \\
& AVG. \# Node & 10 / 20 / 30 & 10 / 20 / 30 & 10 / 20 / 30 & 10 / 20 / 30 & 11 / 20 / 30 & 10 / 20 / 30 & 10 / 20 / 30 & 10 / 20 / 30 \\
& AVG. \# Edge & 55 / 184 / 386 & 55 / 205 / 503 & 49 / 173 / 356 & 36 / 97 / 222 & 31 / 99 / 177 & 46 / 192 / 437 & 32 / 91 / 194 & 47 / 121 / 250 \\
\midrule
& Dataset & \multicolumn{8}{c}{Ours Dataset with Directed Negative Weighted Edge Graphs (Easy)} \\
\multirow{3}{*}{ Directed } & \# Sample & 50 & 50 & 50 & 50 & 50 & 50 & 100 & 100 \\
& AVG. \# Node & 10 & 10 & 10 & 10 & 10 & 10 & 10 & 10 \\
& AVG. \# Edge & 30 & 30 & 28 & 27 & 25 & 22 & 20 & 26 \\
\bottomrule
\end{tabular}
\caption{The summary table of datasets used in the connectivity task including the baseline dataset NLGraph and GraphInstruct. Our dataset is generated for various types of node connections}
\label{all_dataset}
\end{table*}

\section{Example of different descriptions}
\label{description_method}
Here we list the examples of descriptions utilized in the experiment. Examples are listed as node descriptions and edge descriptions on directed or undirected graphs, with or without weights on edges.\\

\textbf{Node Description for Undirected Graph:}
\begin{quote}
\small
\end{quote}

\textbf{Edge Description for Undirected Graph:}
\begin{quote}
\small
G describes an undirected graph among node 0, 1, 2, 3, and 4.\\Node 0 is connected to Node 1.\\Node 1 is connected to Node 2.\\Node 1 is connected to Node 3.\\Node 2 is connected to Node 3.\\Node 3 is connected to Node 4.
\end{quote}

\textbf{Node Description for Directed Graph:}
\begin{quote}
\small
G describes a directed graph among 0, 1, 2, 3, and 4.\\In this graph:\\Node 0 is directed to Node 1.\\Node 1 is directed to Node 2, 3.\\Node 2 is directed to Node 3.\\Node 3 is directed to Node 4.
\end{quote}

\textbf{Edge Description for Directed Graph:}
\begin{quote}
\small
G describes a directed graph among node 0, 1, 2, 3, and 4.\\Node 0 is directed to Node 1.\\Node 1 is directed to Node 2.\\Node 1 is directed to Node 3.\\Node 2 is directed to Node 3.\\Node 3 is directed to Node 4.
\end{quote}

\textbf{Node Description for Undirected Weighted Graph:}
\begin{quote}
\small
G describes an undirected graph among 0, 1, 2, 3, and 4.\\In this graph:\\Node 0 is connected to nodes 1 \\(weight: 8), 2 (weight: 1).\\Node 1 is connected to node 0 \\(weight: 8).\\Node 2 is connected to node 0 \\(weight: 1).
\end{quote}

\textbf{Edge Description for Undirected weighted Graph:}
\begin{quote}
G describes an undirected graph among node 0, 1, 2, 3, and 4.\\Node 0 is connected to Node 1 with weight 8.\\Node 0 is connected to Node 2 with weight 1.
\end{quote}

\textbf{Node Description for Directed weighted Graph:}
\begin{quote}
\small
G describes a directed graph among node 0, 1, 2, 3, and 4.\\In this graph:\\Node 0 is directed to Node 1 (weight: 8), 2 (weight: 1).
\end{quote}

\textbf{Edge Description for Directed weighted Graph:}
\begin{quote}
\small
G describes a directed graph among node 0, 1, 2, 3, and 4.\\Node 0 is directed to Node 1 with \\weight 8.\\Node 0 is directed to Node 2 with \\weight 1.
\end{quote}

\textbf{Knowledge graph Node:}
\begin{quote}
\small
    G describes a knowledge graph among entity: "hairpiece", "wig", "dress", "overdress", "attire", "clothing", and "clothing".
    
Entity "hairpiece" is directed to entity "attire" (relation  hypernym).

Entity "wig" is directed to entity "hairpiece" (relation hypernym).

Entity "dress" is directed to entity "attire" (relation derivationally related form), "dress" (relation verb group), "overdress" (relation also see), and "clothing" (derivationally related form) .

Entity "overdress" is directed to entity "attire" (relation derivationally related form), "dress" (relation verb group).

Entity "attire" is directed to entity "overdress" (relation derivationally related form), "clothing" (relation  hypernym), "dress" (derivationally related form).

Entity "clothing" is directed to entity "dress" (relation derivationally related form).
\end{quote}

\textbf{Knowledge graph Edge:}
\begin{quote}
\small
    G describes a knowledge graph among entity: "hairpiece", "wig", "dress", "overdress", "attire", "clothing", and "clothing".
Entity "hairpiece" is hypernym of entity "attire".

Entity "wig" is hypernym of entity "hairpiece".

Entity "dress" is derivationally related form of entity "attire".

Entity "dress" is verb group of entity "dress".

Entity "dress" is also see of entity "overdress".

Entity "dress" is derivationally related form of entity "clothing".

Entity "overdress" is derivationally related form of entity "attire".

Entity "overdress" is verb group of entity "dress".

Entity "attire" is derivationally related form of entity "overdress".

Entity "attire" is hypernym of entity "clothing".

Entity "attire" is derivationally related form of entity "dress".

Entity "clothing" is derivationally related form of entity "dress".
\end{quote}

\section{Few-shot and CoT examples}
\label{prompts}
Here are examples of how to use few-shot and CoT prompting in graph connectivity and shortest path tasks.

\subsection{Connectivity examples}
\textbf{Few-shot:}
\begin{quote}
\small
Q: Given a directed graph: G describes a directed graph among 0, 1, 2, 3, and 4.\\
In this graph:\\
Node 0 is directed to nodes 1, 3.\\
Node 1 is directed to nodes 2, 0, 4.\\
Node 2 is directed to nodes 3.\\
Node 3 is directed to nodes 4, 0, 1.\\
 Is there a directed path from node 0 to node 3  If the path exist, give "Exist path" the path in the form of "Node \#1 -> Node \#2". Otherwise, give "No path"\\
A: Exist path: 0 -> 3. 
\end{quote}
\textbf{BFS-CoT:}
\begin{quote}
\small
Q: Determine if there is a path between two nodes in the graph. The graph is: G describes an undirected graph among 0, 1, 2, 3, 4, and 5.
In this graph:
Node 0 is connected to node 1.
Node 1 is connected to nodes 0, 2.
Node 2 is connected to nodes 1, 3.
Node 3 is connected to nodes 2, 4.
Node 4 is connected to nodes 3, 5.
Node 5 is connected to node 4.
 The question is: Does a path exist between node 1 and node 5? If a path exists, present the BFS path formatted as "Node \#1 -> Node \#2." If no path is found, state "No path." This problem will be addressed using Breadth-First Search (BFS). Let's think step by step.
A: 1. Node 1 is directely connected with nodes 0, 2.
2. From node 0, we can reach nodes 1.
3. From node 2, we can reach nodes 1, 3.
4. From node 3, we can reach nodes 2, 4.
5. From node 4, we can reach nodes 3, 5.
6. From node 5, we have reached our target.
So a path from node 1 to node 5 does exist. The BFS path taken in this search is:
Node 1 -> Node 2 -> Node 3 -> Node 4 -> Node 5.
\end{quote}

\textbf{Dijkstra-CoT:}
\begin{quote}
\small
Q: Given a directed graph: G describes a directed graph among 0, 1, 2, 3, and 4.\\
In this graph:\\
Node 0 is directed to nodes 1 (weight: 6), 4 (weight: 6), 3 (weight: 6), 2 (weight: 1).\\
Node 1 is directed to nodes 2 (weight: 4), 4 (weight: 3).\\
Node 2 is directed to nodes 3 (weight: 7), 1 (weight: 4), 0 (weight: 5).\\
Node 3 is directed to nodes 4 (weight: 6), 0 (weight: 6).\\
Node 4 is directed to nodes 1 (weight: 6), 3 (weight: 9), 0 (weight: 5).\\
 Does a path exist from node 4 to node 2? If so, provide the shortest path formatted as "Node \#1 -> Node \#2." If not, state "No path.". Let's think step by step using Dijkstra's Algorithm.\\
A: To determine if there is a path from node 4 to node 2 and find the shortest path if it exists, we can use Dijkstra's Algorithm. Let's apply the algorithm step by step:\\
\\
Initialization\\
- Start with node 4.\\
- Set the distance to node 4 (the starting node) to 0 and to all other nodes to infinity.\\
- Keep a priority queue to select the node with the smallest tentative distance that hasn't been permanently set yet.\\
- Mark all nodes as unvisited.\\
\\
Step by Step Process\\
1. Select node 4 (smallest distance in the priority queue). From node 4, we can reach nodes 1, 3 and 0. We update the distances to nodes 1, 3 and 0.\\
  - Distance to node 1 (from node 4) is 6, which is better than the previous, update the priority queue.\\
  - Distance to node 3 (from node 4) is 9, which is better than the previous, update the priority queue.\\
  - Distance to node 0 (from node 4) is 5, which is better than the previous, update the priority queue.\\
\\
2. Select node 0 (next smallest distance in the priority queue). From node 0, we can reach nodes 1, 4, 3 and 2. However, nod and 4 has already been selected. We update the distances to nodes 1, 3 and 2.\\
  - Distance to node 1 (from node 0) is 11, which is not better than the previous, and will not update the priority queue.\\
  - Distance to node 3 (from node 0) is 11, which is not better than the previous, and will not update the priority queue.\\
  - Distance to node 2 (from node 0) is 6, which is better than the previous, update the priority queue.\\
\\
3. Select node 1 (next smallest distance in the priority queue). From node 1, we can reach nodes 2 and 4. However, node 4 has already been selected. We update the distance to node 2.\\
  - Distance to node 2 (from node 1) is 10, which is not better than the previous, and will not update the priority queue.\\
\\
4. Select node 2 (next smallest distance in the priority queue). From node 2, we can reach nodes 3, 1 and 0. However, nodes 1 and 0 have already been selected. We update the distance to node 3.\\
  - Distance to node 3 (from node 2) is 13, which is not better than the previous, and will not update the priority queue.\\
\\
5. Select node 3 (next smallest distance in the priority queue). From node 3, we can reach nodes 4 and 0. However, nodes 4 and 0 have already been selected. We have nothing to update.\\
\\
Conclusion\\
A path exists from node 4 to node 2.\\
Answer: Node 4 -> Node 0 -> Node 2.\\
\end{quote}

\section{Graph Connectivity task}
\subsection{Dataset}
\label{graph_connect_datasets}

Table \ref{all_dataset} shows the dataset information for all datasets we used. We separated the dataset according to the connectivity type to ensure clear display of whether or not the dataset is uniformly distributed. The table helps illustrate that our dataset aims to balance the distribution on connectivity type as uniform as possible (with 50 samples for most types), as well as the balance of positive and negative cases (250 samples for both connective and non-connective cases).

\subsection{Results}
\label{graph_connect_results}

Table \ref{baseline_all} shows the zero-shot accuracy result of baseline datasets. The result is separated by connectivity type in columns. However, due to the variability of distribution, significant numbers of grids remain empty. Table \ref{baseline_all_pathundirected} and Table \ref{baseline_all_pathdirected} are novel evaluations of undirected and directed baseline datasets with ACC and $\mathrm{F_{acc}}$.
% \input{latex/tables/acc_table}

% \subsection{Failed cases}

\section{Shortest-path task}
\subsection{Experimental setup}
\label{sec:shortest-path-experiment_setting}
We study the shortest-path problem using the Easy datasets from the unweighted graphs as mentioned in Section~\ref{connectivity}. For the weighted graphs, we applied similar strategies that were used in undirected graph generations to generate the directed and undirected graph datasets.
The directed graph datasets include two types, whether there are negative edges in the graphs. Appendix~\ref{graph_connect_datasets} Table~\ref{all_dataset} shows the details. The graph structure descriptions are shown in Appendix~\ref{description_method}

\subsection{Result}

Table \ref{strategy2} records the shortest path accuracy on various prompting methods. Weighted graph in this table only have positive weights.

\subsection{Analysis of other factors}
\label{analysis}
% \begin{table}[]
%   \centering
%   \small
% \begin{tabular}{cccccc}
% \multirow{2}{*}{ Strategies } & \multirow{2}{*}{ Prompt } & \multicolumn{2}{c}{edge} & \multicolumn{2}{c}{node} \\
% & & Acc & F_acc & Acc & F_acc \\
% \multirow{2}{*}{ No strategies } & zero-shot & 0.81 & 0.87 & 0.87 & 0.83 \\
% & few-shot & & & & \\
% \multirow{2}{*}{ DFS } & zero-shot & 0.936 & 0.972 & 0.95 & 0.984 \\
% & CoT & 0.958 & 0.9 & 0.95 & 0.952 \\
% \multirow{2}{*}{ BFS } & zero-shot & 0.938 & 0.912 & 0.9 & 0.932 \\
% & CoT & 0.926 & 0.94 & 0.922 & 0.972 \\
% \end{tabular}
%     \caption{Caption}
%     \label{strategy_1}
% \end{table}

% \begin{table}[]
%   \centering
%   \small
%       \caption{Results for graph connectivity with algorithm prompts}
% \begin{tabular}{@{\hskip 0in}c@{\hskip 0in}c|cccc}
% \toprule
% \multirow{2}{*}{ Algorithms } & \multirow{2}{*}{ Prompt } & \multicolumn{2}{c}{Node} & \multicolumn{2}{c}{Edge} \\
% & & ACC & $\mathrm{F_{acc}}$ & ACC & $\mathrm{F_{acc}}$ \\
% \midrule
% \multirow{2}{*}{ No Algorithms } & 0-shot & 0.81 & 0.78 & 0.87 & 0.83 \\
% & few-shot & 0.82 & 0.95 & 0.78 & 0.89 \\
% \multirow{2}{*}{ DFS } & 0-shot & 0.95 & 0.98 & 0.94 & 0.97 \\
% & CoT & 0.95 & 0.95 & 0.96 & 0.90 \\
% \multirow{2}{*}{ BFS } & 0-shot & 0.90 & 0.93 & 0.94 & 0.91 \\
% & CoT & 0.92 & 0.97 & 0.93 & 0.94 \\
% \bottomrule
% \end{tabular}
%     \label{strategy_1}
%     \vspace{-0.2in}
% \end{table}

\begin{table*}[]
    \centering
    \small
    \caption{Algorithm CoT applied in the graph connectivity and shortest path}
    \begin{tabular}{c|c|cccccc}
\toprule
\multicolumn{8}{c}{Connectivity task ($\mathrm{F_{acc}}$)} \\
Dataset & prompt & k-hop, 1$\leq$k$\leq$2 & k-hop, 3$\leq$k$\leq$4 & 5-hop & Singleton & I.C. & AVG. \\\midrule
\multirow{3}{*}{ Undirected } & 0-shot & 0.93 & 0.93 & 0.94 & 1.00 & 0.53 & 0.78 \\
& few-shot & 0.92 & 0.93 & 0.96 & 1.00 & 0.87 & 0.92 \\
& BFS-CoT & 0.95 & 0.98 & 1.00 & 1.00 & 0.88 & 0.93 \\\midrule
\multicolumn{8}{c}{Shortest path (ACC)} \\\midrule
\multirow{3}{*}{ undirected } & 0-shot & 0.88 & 0.87 & 0.88 & 1.00 & 0.46 & 0.72 \\
& few-shot & 0.91 & 0.90 & 0.78 & 1.00 & 0.52 & 0.75 \\
& Dijkstra-CoT & 0.96 & 0.94 & 0.86 & 1.00 & 0.70 & 0.84 \\\midrule
\multirow{3}{*}{ weighted undirected } & 0-shot & 0.80 & 0.52 & 0.54 & 0.98 & 0.47 & 0.60 \\
& few-shot & 0.75 & 0.58 & 0.48 & 0.92 & 0.39 & 0.56 \\
& Dijkstra-CoT & 0.81 & 0.65 & 0.58 & 0.84 & 0.53 & 0.65 \\\bottomrule
\end{tabular}
    
    \label{strategy_1}
    \vspace{-0.1in}
\end{table*}

% \begin{table*}[]
%     \centering
%     \tiny
%     \begin{tabular}{ccccccccccccccc}
% \multirow{3}{*}{ Naming } & \multirow{3}{*}{ Des. } & \multicolumn{11}{c}{Sub-datasets for undirected graphs} & \multirow{3}{*}{ AVg. ACC } & \multirow{3}{*}{ AVg. $\mathrm{F_{ACC}}$ } \\
% & & \multicolumn{3}{c}{k-hop, 1<k<2} & \multicolumn{3}{c}{k-hop, 3<k<4} & \multicolumn{3}{c}{5-hop} & attached & Isolated C. & & \\
% & & ACC & $\mathrm{F_{ACC}}$ & PCR & ACC & $\mathrm{F_{ACC}}$ & PCR & ACC & $\mathrm{F_{ACC}}$ & PCR &$\mathrm{F_{ACC}}$ & $\mathrm{F_{ACC}}$ & & \\
% \multirow{2}{*}{ Ordered ID } & Node & 1 & 0.93 & 0.9215 & 1 & 0.93 & 0.919 & 0.98 & 0.92 & 0.893 & 1 & 0.705 & 0.88 & 0.846 \\
% & Edge & 1 & 0.93 & 0.8845 & 1 & 0.9 & 0.8845 & 0.98 & 0.88 & 0.858 & 1 & 0.815 & 0.924 & 0.88 \\
% \multirow{2}{*}{ Random ID } & Node & 1 & 0.81 & 0.81 & 0.99 & 0.84 & 0.812 & 0.94 & 0.88 & 0.859 & 1 & 0.595 & 0.828 & 0.754 \\
% & Edge & 1 & 0.89 & 0.875 & 0.99 & 0.88 & 0.852 & 0.88 & 0.68 & 0.64 & 1 & 0.735 & 0.88 & 0.816 \\
% \multirow{2}{*}{ Random  characters } & Node & 1 & 0.823 & 0.8235 & 1 & 0.86 & 0.8575 & 0.98 & 0.94 & 0.927 & 1 & 0.62 & 0.846 & 0.78 \\
% & Edge & 1 & 0.88 & 0.86 & 0.98 & 0.87 & 0.8565 & 0.88 & 0.82 & 0.785 & 1 & 0.696 & 0.8624 & 0.8104 \\
% \end{tabular}
%     \caption{Results for different node named methods}
%     \label{named_results}
% \end{table*}

\begin{table*}[]
    \centering
    \small
        \caption{Results for different node ID naming methods}
    \setlength{\tabcolsep}{5pt}
\resizebox{\textwidth}{!}{% 
\begin{tabular}{c@{\hskip 0in}c|ccccccccccccc}
\toprule
\multirow{2}{*}{ Naming } & \multirow{2}{*}{ Des. } & \multicolumn{3}{c}{k-hops, 1$\leq$k$\leq$2} & \multicolumn{3}{c}{k-hop, 3$\leq$k$\leq$4} & \multicolumn{3}{c}{5-hop} & Singleton & Isolated C. & \multirow{2}{*}{ AVG. ACC } & \multirow{2}{*}{ AVG. $\mathrm{F_{acc}}$ } \\
& & ACC & $\mathrm{F_{acc}}$ & PCR & ACC & $\mathrm{F_{acc}}$ & PCR & ACC & $\mathrm{F_{acc}}$ & PCR & $\mathrm{F_{acc}}$ & $\mathrm{F_{acc}}$ & & \\
\midrule
\multirow{2}{*}{ Ordered ID } & Node & 1.00 & 0.93 & 0.99 & 1.00 & 0.93 & 0.99 & 1.00 & 0.94 & 0.97 & 1.00 & 0.53 & 0.81 & 0.78 \\
& Edge & 1.00 & 0.93 & 0.98 & 1.00 & 0.90 & 0.98 & 0.98 & 0.88 & 0.97 & 0.98 & 0.69 & 0.87 & 0.83 \\
\multirow{2}{*}{ Random ID } & Node & 1.00 & 0.81 & 1.00 & 1.00 & 0.85 & 1.00 & 1.00 & 0.92 & 0.97 & 1.00 & 0.41 & 0.77 & 0.69 \\
& Edge & 1.00 & 0.89 & 0.98 & 0.99 & 0.88 & 0.97 & 0.96 & 0.70 & 0.94 & 1.00 & 0.59 & 0.83 & 0.76 \\
\multirow{2}{*}{ Random characters } & Node & 1.00 & 0.83 & 0.99 & 1.00 & 0.86 & 1.00 & 1.00 & 0.94 & 0.99 & 0.98 & 0.43 & 0.77 & 0.70 \\
& Edge & 1.00 & 0.88 & 0.98 & 0.99 & 0.88 & 0.98 & 0.94 & 0.88 & 0.96 & 0.98 & 0.55 & 0.81 & 0.76 \\
\bottomrule
\end{tabular}
}
    \label{named_results}
    \vspace{-0.1in}
\end{table*}

\subsubsection{Impact of the algorithm prompts}
\label{strategies}
In-context learning approaches, including Chain-of-Thought (CoT)~\cite{wei2022chain} and zero-Chain-of-Thought (0-CoT)~\cite{kojima2022large}, have been widely utilized in LLMs to enhance their reasoning capabilities. Meanwhile, specifically in graph-related tasks, previous works combined the prompts with the graph algorithms. However, they do not demonstrate consistent improvement~\cite{wang2024can}. In this subsection, we revisit these approaches in detail.

We consider several graph algorithms in the experiments. For the graph connectivity task, we focus on the Breadth-First Search (BFS) and we employ the Dijkstra algorithms to soleve the shortest path problem. We utilize Node descriptions to search the connectivity and shortest pathes in Easy setting by GPT-4. The prompts examples are shown in Appendix~\ref{prompts}. The results are detailed in Table \ref{strategy_1}.

The observations can be summarized as follows: (1) In the connectivity task, few-shot examples help LLMs recognize isolated components. This is because few-shot examples enable the LLMs to correctly output 'No connection' when they do not find a connected path. (2) In the shortest path cases, few-shot examples do not consistently lead to better performance. However, performance improves when the Dijkstra-CoT method is applied. This suggests that while LLMs may use multiple strategies to make decisions, but a specific algorithm can guide them toward a unique solution.

\subsubsection{The influence of node names}
\label{naming}
\citet{fatemi2023talk} suggest that different naming methods for graphs can yield varied results. This variation is attributed to the graph node IDs occupying the same space as the pre-trained data of LLMs. Thus, we further evaluated the impact of naming conventions on nodes for the connectivity task. Table~\ref{named_results} summarizes the results for GPT-4 on the Easy subset of the undirected graph dataset. "Ordered ID" refers to nodes named sequentially as "1, 2, 3, ...", "Random ID" denotes nodes named using random numbers up to 10,000, and "Random character" represents nodes named with random five-character strings. The results indicate that naming nodes in sequential order, a common practice in graph descriptions, may enhance LLM performance. This suggests that LLMs could leverage some form of memory recognition to predict connectivity more effectively and thus achieve higher performance.

\section{Knowledge graph}
\subsection{Dataset}
We used WN18RR \cite{shang2019end} as the base dataset, which provides both ID names and Entity names. The ID names consist of strings of random numbers, and Entity names are used as specific and meaningful identifiers. From its training set, we randomly selected 150 subgraphs based on ego graphs with a depth of 3. Within each subgraph, we identified two nodes with the longest paths and segmented the paths into $k'$-hops. This strategy allowed us to generate $k'$ question-answer pairs, ranging from 1-hop to $k'$-hop.
% Table~\ref{app_datasets} shows the details of our sampled dataset. We take both Node List and Edge List in the experiment. The description examples can be found in Appendix~\ref{description_method}.

Table \ref{app_datasets} contains information about knowledge graph dataset, Including number of samples, average number of nodes, average number of edges in all connectivity types.

\section{K-hops influence on the connectivity task}

In Section~\ref{connectivity}, we have demonstrated that performance in the graph connectivity task is closely related to the number of nodes and k-hops in a graph. However, it is important to note that smaller graphs inherently support shorter paths. To fairly assess the impact of k-hops on different graph sizes, we further evaluate the relations between k-hop and graph density. 

\begin{figure*}[htbp]  
  \centering  
  % 第一张图  
  \begin{minipage}[t]{0.48\linewidth}  
    \centering  
    \includegraphics[width=\linewidth]{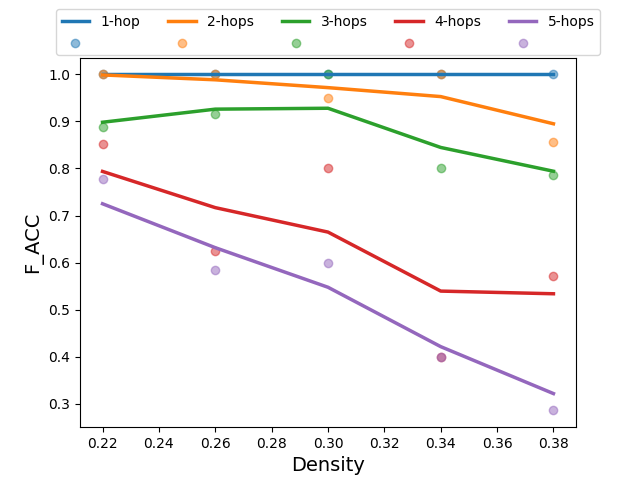}  
    \caption{Accuracy of K-hops across varying graph densities (Node List)}  
    \label{k}  
  \end{minipage}  
  \hfill   
  \begin{minipage}[t]{0.48\linewidth}  
    \centering  
    \includegraphics[width=\linewidth]{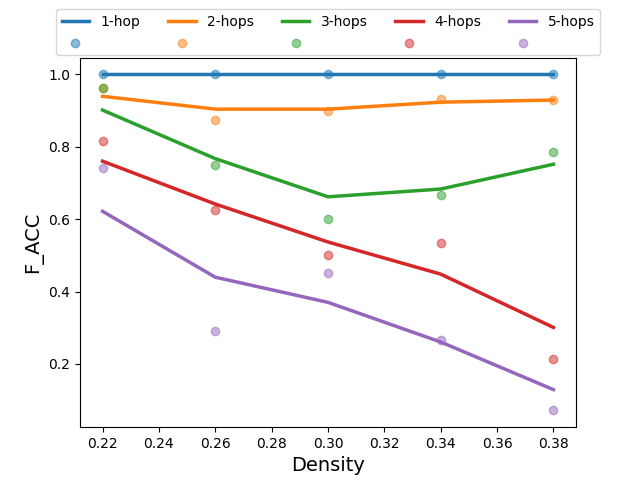}  
    \caption{Accuracy of K-hops across varying graph densities (Edge List)}  
    \label{fig:enter-label}  
  \end{minipage}  
\end{figure*}

We create a subset with 100 undirected graphs where the graph node number is 16 - 36 and the density is in the range of (0.2,0.4) and evaluate them by Node and Edge List descriptions. The results are shown in Figure~\ref{k}.
% \begin{figure}
%     \centering
%     \includegraphics[width=0.48\textwidth]{latex/figs/k-hop.png}
%     \caption{Accuracy of K-hops across varying graph densities (Node List)}
%     \label{k}
% \end{figure}
% \begin{figure}
%     \centering
%     \includegraphics[width=0.48\textwidth]{latex/figs/adjfigure.png}
%     \caption{Accuracy of K-hops across varying graph densities (Edge List)}
%     \label{fig:enter-label}
% \end{figure}

% \subsection{Density influence on the connectivity tasks}
% The results indicate that 1-hop cases maintain 100\% \jt{the accuracy is the normal one or the one we proposed??} 
The results indicate that 1-hop cases maintain a very high accuracy regardless of graph density, while 2-hop and 3-hop cases show a slight accuracy decrease. In contrast, 4-hop and 5-hop cases exhibit high accuracy only in sparse graphs but significantly decline when graph density approaches 0.38. This suggests that LLMs become confused as the graph complexity increases.

Comparing the Node List and Edge List descriptions, it is observed that the Node List exhibits a smaller reduction in performance compared to the Edge List. This suggests that the Node List may be more effective in describing complex graphs.

\section{The explanations for the reasoning ability of LLMs in $K$-hops}

\subsection{Traing on GPT-2}
\label{sec:training_details}
We adopt the transformer reasoning framework proposed in~\cite{alpine_24} to train a $k$-hop-specific reasoning model. Specifically, we simplify the edge list and node list by converting them into node symbols, as shown in Table~\ref{tab:des}.

\begin{table*}[]
\caption{The simplified description forms of graphs}
\label{tab:des}
\centering
\begin{tabular}{l|p{15em}|p{15em}}
\toprule
Graph & Node list                                                                                                                     & Edge list                                                                                                                       \\
\midrule
    \raisebox{-0.5cm}{\includegraphics[width=0.07\linewidth]{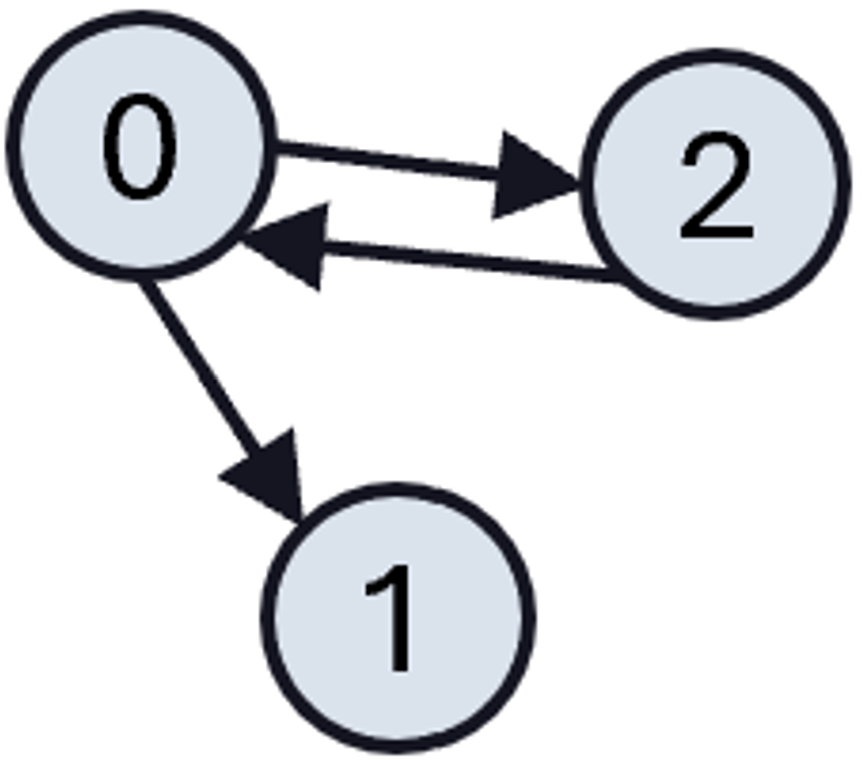}}

      & \textless{}START\_Q\textgreater 0: 1, 2; 2: 0 between 2, 1 \textless{}END\_Q\textgreater 2, 0, 1 \textless{}END\textgreater{} & \textless{}START\_Q\textgreater{} 0 1 | 0 2 | 2 0 between 2, 1 \textless{}END\_Q\textgreater 2, 0, 1 \textless{}END\textgreater{}\\
      \bottomrule
\end{tabular}
\end{table*}

We select different scales of the GPT-2 model, as shown in Table~\ref{tab:model_scale1}. During training, we sequentially predict the nodes on the shortest path using cross-entropy loss, following the approach outlined in~\cite{alpine_24}. The models were trained for 20,000 epochs on a single H100 GPU, starting with a learning rate of 1e-4, which is reduced to 1e-5 after 20,000 epochs. We apply a dropout rate of 0.2 and saved the best-performing model based on the validation set. During testing, we evaluate the loss across all predicted tokens.

\subsection{Finetune on LLama}
\label{sec:finetune}
\begin{figure}
    \centering
    \includegraphics[width=0.5\linewidth]{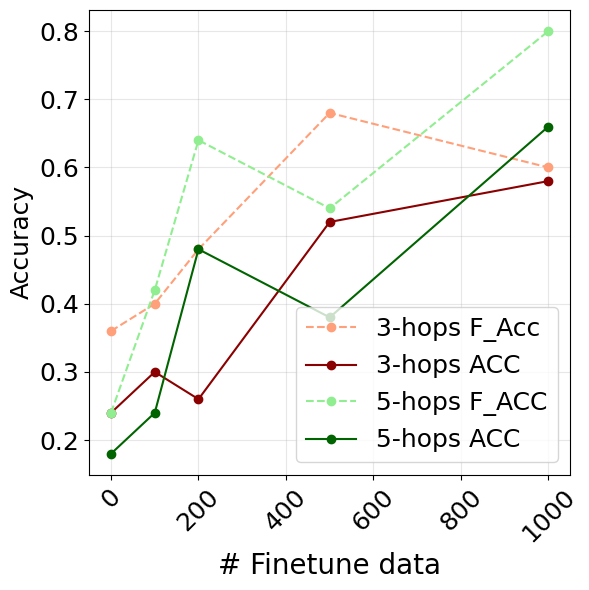}
    \caption{The effect of fine-tuning on the LLama.}
    \label{fig:finetune}
\end{figure}

Furthermore, we evaluate the impact of data on fine-tuning. The fine-tuning data is derived from a specifically designed dataset with standard graph descriptions, as illustrated in Figure~\ref{fig-descript}. The test data aligns with the dataset described in~\ref{all_dataset}. We use Llama3.2-3B as the backbone model and apply LoRA for fine-tuning to enhance shortest-path reasoning. The results are presented in Figure~\ref{fig:finetune}.

We find that only a little finetuning data can make the model have better performance. Although we are limited by the computation resource, we believe more data can drive model perform well, which is align with our observation on the training setting.

\begin{table}
\small
\centering

\caption{The version of GPT-2 models}
\begin{tabular}{lllll}
\toprule
            & \# Params & \multicolumn{1}{l}{\# head} & \multicolumn{1}{l}{\# layer} & \multicolumn{1}{l}{\# emb} \\
            \midrule
Baby-GPT2   & 80M        & 6                            & 6                             & 384                         \\
Small-GPT2  & 124M       & 12                           & 12                            & 768                         \\
Medium-GPT2 & 350M       & 24                           & 16                            & 1024                   \\
\bottomrule
\end{tabular}
\label{tab:model_scale1}
\end{table}

\section{Reasoning process}
\label{sec:app_reasoning_process}
We drew inspiration from syntax analysis in language models~\cite{bert_vis} to the reasoning analysis.
In the graph description, \textless{}S\textgreater\ and \textless{}T\textgreater\ are designated as the source and target nodes, highlighted in green. \textless{}PATH\textgreater$_i$ in gray, denotes the $i$-th node on the path from the source node to the target node, while \textless{}O\_N\textgreater\ represents other nodes, shown in yellow. In the question, the source and target nodes are labeled as \textless{}SQ\textgreater\ and \textless{}TQ\textgreater\ in blue, respectively. When the models predict the $i+1$-th nodes, they require previous information, denoted as \textless{}PRED$_i$\textgreater\ in red. 

\begin{figure*}
    \centering
    \includegraphics[width=0.9\linewidth]{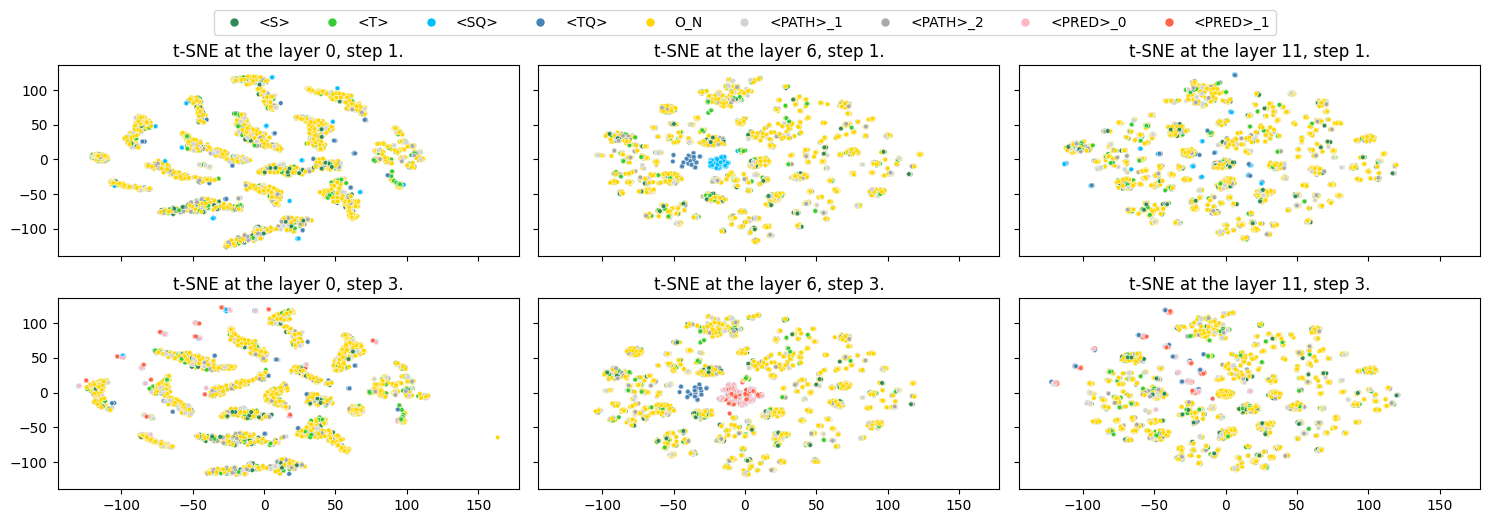}
    \caption{The t-SNE results on the Node List.}
    \label{fig:tsne_adj}
\end{figure*}

\begin{figure*}
    \centering
    \includegraphics[width=0.9\linewidth]{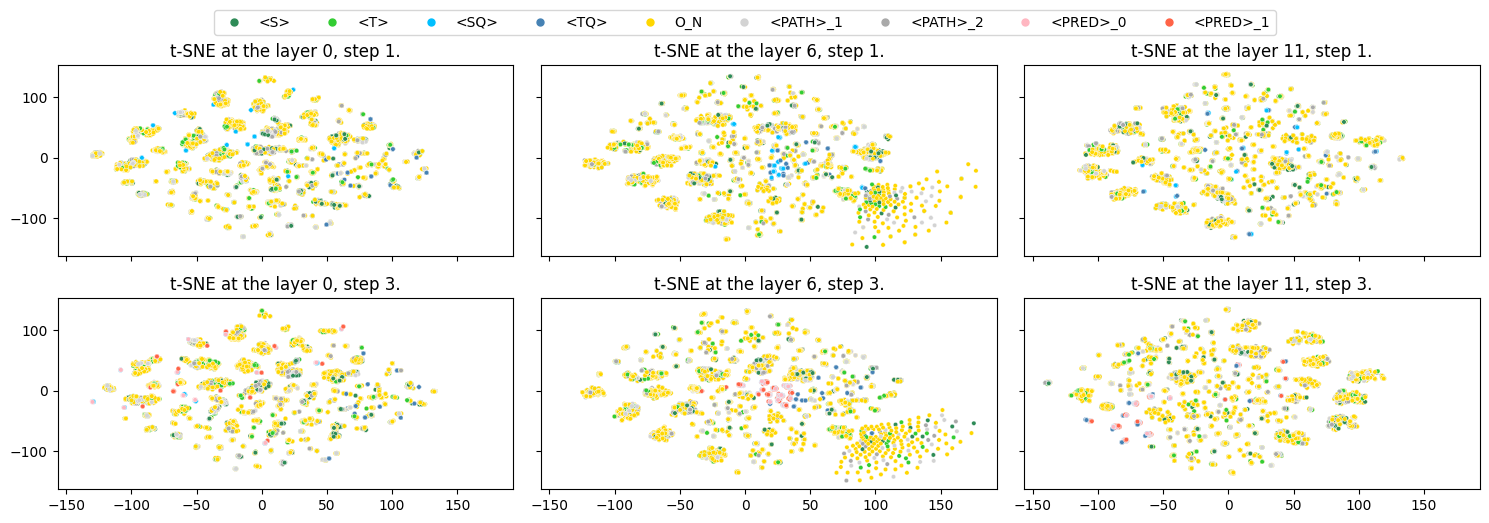}
    \caption{The t-SNE results on the Edge List.}
    \label{fig:tsne_edge}
\end{figure*}

Figure~\ref{fig:tsne_adj} and Figure~\ref{fig:tsne_edge} provides more detail information when the GPT-2 do the reasoning with different graph descriptions.

\section{Failed cases}
\label{translation_failed}
In this section, we will list some failed cases. We mark the added edges in \wrongs{Red} and ignored edges in \ignore{Green}.
\subsection{Translation for Edge List to Node List}
\begin{quote}
\small
\textbf{Question:}
Your task is giving the neighbors of each node.G describes an undirected graph among node 0, 1, 2, 3, 4, 5, 6, 7, 8, 9, 10, 11, and 12.

Node 0 is connected to Node 1.
Node 0 is connected to Node 5.
Node 0 is connected to Node 9.
Node 0 is connected to Node 12.
Node 0 is connected to Node 3.
Node 0 is connected to Node 10.
Node 0 is connected to Node 8.
Node 0 is connected to Node 11.
Node 0 is connected to Node 7.

Node 1 is connected to Node 2.
Node 1 is connected to Node 4.
Node 1 is connected to Node 3.
Node 1 is connected to Node 12.
Node 1 is connected to Node 9.
Node 1 is connected to Node 11.
Node 1 is connected to Node 10.
Node 1 is connected to Node 5.
Node 1 is connected to Node 6.

Node 2 is connected to Node 3.
Node 2 is connected to Node 4.
Node 2 is connected to Node 6.
Node 2 is connected to Node 10.
Node 2 is connected to Node 9.
Node 2 is connected to Node 12.
Node 2 is connected to Node 7.
Node 2 is connected to Node 11.

Node 3 is connected to Node 4.
Node 3 is connected to Node 11.
Node 3 is connected to Node 5.
Node 3 is connected to Node 10.
Node 3 is connected to Node 12.
Node 3 is connected to Node 8.

Node 4 is connected to Node 5.
Node 4 is connected to Node 9.
Node 4 is connected to Node 8.
Node 4 is connected to Node 10.
Node 4 is connected to Node 6.
Node 4 is connected to Node 11.
Node 4 is connected to Node 7.

Node 5 is connected to Node 6.
Node 5 is connected to Node 9.
Node 5 is connected to Node 10.
Node 5 is connected to Node 7.
Node 5 is connected to Node 8.

Node 6 is connected to Node 7.
Node 6 is connected to Node 9.
Node 6 is connected to Node 12.
Node 6 is connected to Node 8.

Node 7 is connected to Node 8.
Node 7 is connected to Node 10.
Node 7 is connected to Node 9.
Node 7 is connected to Node 12.

Node 8 is connected to Node 9.
Node 8 is connected to Node 12.
Node 8 is connected to Node 10.
Node 8 is connected to Node 11.

Node 9 is connected to Node 10.
Node 9 is connected to Node 11.
Node 9 is connected to Node 12.

Node 10 is connected to Node 11.
Node 10 is connected to Node 12.
Node 11 is connected to Node 12.
For each node, the neighboring nodes are listed as follows: "Node \#1: []
 Node \#2: []"
 
 \textbf{Answer:}
Given the connections in the undirected graph, the neighbors of each node can be summarized as follows:

- Node 0: [1, 5, 9, 12, 3, 10, 8, 11, 7]
- Node 1: [0, 2, 4, 3, 12, 9, 11, 10, 5, 6]
- Node 2: [1, 3, 4, 6, 10, 9, 12, 7, 11]
- Node 3: [0, 1, 2, 4, 11, 5, 10, 12, 8]
- Node 4: [1, 2, 3, 5, 9, 8, 10, 6, 11, 7]
- Node 5: [0, 1, 3, 4, 6, 9, 10, 7, 8]
- Node 6: [1, 2, 4, 5, 7, 9, 12, 8]
- Node 7: [0, 2, 4, 5, 6, 8, 10, 9, 12]
- Node 8: [0, 3, 4, 5, 6, 7, 9, 12, 10, 11]
- Node 9: [0, 1, 2, 4, 5, 6, 7, 8, 10, 11, 12]
- Node 10: [0, 1, 2, 3, 4, 5, 7, 8, 9, 11, 12]
- Node 11: [0, 1, 2, 3, 4, 8, 9, 10, 12]
- Node 12: [0, 1, 2, 3, \wrongs{4, 5,} 6, 7, 8, 9, 10, 11]
\end{quote}

\subsection{Translation for Node List to Edge List}
\begin{quote}
\small
\textbf{Question:}
Your task is to identify all the edges in a graph and represent them in sparse form.G describes an undirected graph among 0, 1, 2, 3, 4, 5, 6, 7, 8, 9, 10, 11, 12, and 13.
In this graph:

Node 0 is connected to nodes 1, \ignore{11}, 2.

Node 1 is connected to nodes 0, 2, 3, 4, 9, 11, 7.

Node 2 is connected to nodes 0, 1, 3, 9, 8, 10.

Node 3 is connected to nodes 1, 2, 4, 8, 6, 13, 12.

Node 4 is connected to nodes 1, 3, 5, 10, \ignore{9}.

Node 5 is connected to nodes 4, 6, 9, 7.

Node 6 is connected to nodes 3, 5, 7.

Node 7 is connected to nodes 1, 5, 6, 8, 12, 9, 11.

Node 8 is connected to nodes 2, 3, 7, 9, 12, 10.

Node 9 is connected to nodes 1, 2, 4, 5, 7, 8, 10.

Node 10 is connected to nodes 2, 4, 8, 9, 11.

Node 11 is connected to nodes 0, 1, 7, 10, 12.

Node 12 is connected to nodes 3, 7, 8, 11, 13.

Node 13 is connected to nodes 3, 12.

The edges should be list as: (Node \#1, Node \#2), ...

\textbf{Answer:}
To represent the edges of the given undirected graph in sparse form, we'll list each connection between nodes as a tuple `(Node \#1, Node \#2)`. Since the graph is undirected, each edge is bidirectional, meaning if `(Node A, Node B)` is listed, `(Node B, Node A)` is not listed again to avoid redundancy. We'll go through each node's connections as provided, ensuring to list each edge only once.

Here are the edges in sparse form:

```
(0, 1), (0, 11), (0, 2),
(1, 2), (1, 3), (1, 4), (1, 9), (1, 7),
(2, 3), (2, 9), (2, 8), (2, 10),
(3, 4), (3, 8), (3, 6), (3, 13), (3, 12),
(4, 5), (4, 10),
(5, 6), (5, 9), (5, 7),
(6, 7),
(7, 8), (7, 12), (7, 9), (7, 11),
(8, 9), (8, 12), (8, 10),
(9, 10),
(10, 11),
(11, 12),
(12, 13)
```

This list includes all unique edges in the graph, ensuring that each connection is represented once, respecting the undirected nature of the graph.
\end{quote}

\end{document}